\documentclass[9pt,twocolumn,twoside]{include/osajnl}
\journal{include/ol} 
\setboolean{shortarticle}{false}

\title{Rapid On-Robot Learning for Dynamic Manipulation Skills: Robot Juggling}

\author[*,1]{Taeyoon Lee}
\author[1]{Chunpeng Wang}
\author[1,2]{Christopher G. Atkeson}
\author[1]{Alfred A. Rizzi}
\author[1]{Nicolas Rojas}

\affil[1]{Robotics and AI Institute (RAI), Cambridge MA, USA}
\affil[2]{Robotics Institute, Carnegie Mellon University, Pittsburgh PA, USA}
\affil[*]{Corresponding author: \href{mailto:tlee@rai-inst.com}{tlee@rai-inst.com}}

\usepackage[]{units} 
\usepackage{xspace} 
\usepackage{hyperref} 
\usepackage{authblk}  
\usepackage[table]{xcolor}
\usepackage{booktabs}
\usepackage{tabularx}
\usepackage{tabularray}
\UseTblrLibrary{booktabs}
\usepackage{xcolor}
\usepackage[nonumberlist, nogroupskip, section=section, numberedsection=autolabel]{glossaries}
\glsdisablehyper
\newacronym{rl}{RL}{Reinforcement Learning}
\newacronym{umv}{UMV}{Ultra Mobility Vehicle}  
\newacronym{cmdp}{CMDP}{Constrained Markov Decision Process}
\newacronym{mdp}{MDP}{Markov Decision Process}
\newacronym{rsi}{RSI}{Reference State Initialization}
\newacronym{ppo}{PPO}{Proximal Policy Optimization}
\newacronym{mlp}{MLPs}{Multi-Layer Perceptrons}
\newacronym{rlhf}{RLHF}{Reinforcement Learning from Human Feedback}

\usepackage{newfloat}
\DeclareFloatingEnvironment[
    fileext=lom,      
    listname={List of Movies}, 
    name=Movie,       
    placement=tbhp,   
]{movie}

\begin{document} 

\begin{abstract} \bfseries \boldmath \sffamily
\vspace*{6pt}
We present an online learning framework that enables a bimanual robot to acquire diverse juggling patterns directly on physical hardware within minutes, even with a significant sim-to-real gap. One of the most important lessons from this work is that a model, even when far from reality, can be extremely useful for learning. This motivates a central philosophy of our approach: learning should build upon the robot’s current knowledge rather than replace it. Our regularized memory-based learning puts this principle into practice by learning a local model from accumulated experience while retaining the global prior model to extrapolate where experience is sparse. This enables efficient and stable online learning from each new experience without resorting to uninformed exploration over a vast space of possible behaviors. Equally important to continual on-robot learning is safety, allowing the robot to repeatedly practice and improve in the real world. We construct a mutually reachable set that allows safe transitions between successive throws and catches, without driving either arm into a state from which its next action would require violating the robot’s joint or actuator limits. Together, these ideas enable a bimanual robot with multi-fingered hands and onboard vision to safely learn and compose five canonical three-ball juggling patterns—including cascade, tennis, half-shower, shower, and box—within less than 5 minutes of real-world interaction. More broadly, this work points toward robots that build upon imperfect prior knowledge and continually refine their behavior through their own real-world experience.

\vspace*{6pt}
\end{abstract}

\maketitle

\section{Introduction}
What can robots do to overcome a significant gap between simulation and reality (the sim-to-real gap)? Prior knowledge derived from simulation, 
analytical models, 
or large-scale demonstration datasets 
often provides a strong starting point for executing robot behavior in the physical world, sometimes enabling impressive robust (often referred to as ``zero-shot'') performance with little or no additional real-world learning. However, no model or prior experience captures every detail of reality at deployment, and particularly for high-performance manipulation, even small discrepancies can significantly affect task performance \cite{yang2024tube}. This is especially true when motions are fast, contacts are intermittent, and individual actions must be precisely coordinated in space and time. When zero-shot execution falls short, how should a robot system be engineered to improve beyond its initial performance?

We study this question through robot juggling, a demanding example of high-performance dynamic manipulation. Each throw must deliver a ball to the right place at the right time for a future catch, with little opportunity for correction once the ball is in flight. Small errors affect subsequent catches and throws and can quickly accumulate over repeated exchanges. In particular, we consider juggling with multi-fingered robotic hands, where the ball can be grasped differently within the hand after every catch, and subtle differences in how the fingers interact with and release the ball can significantly change the next throw \cite{juzaju2012two}. Together with variations in the robot's actuation, these factors make the outcome of each throw difficult to model and predict accurately.

\begin{movie*}[t]
    \centering
    \href{https://youtu.be/tAPvN-tQpX0?si=rphZ-NoF9fS-n4kb}{%
    \includegraphics[width=\linewidth]{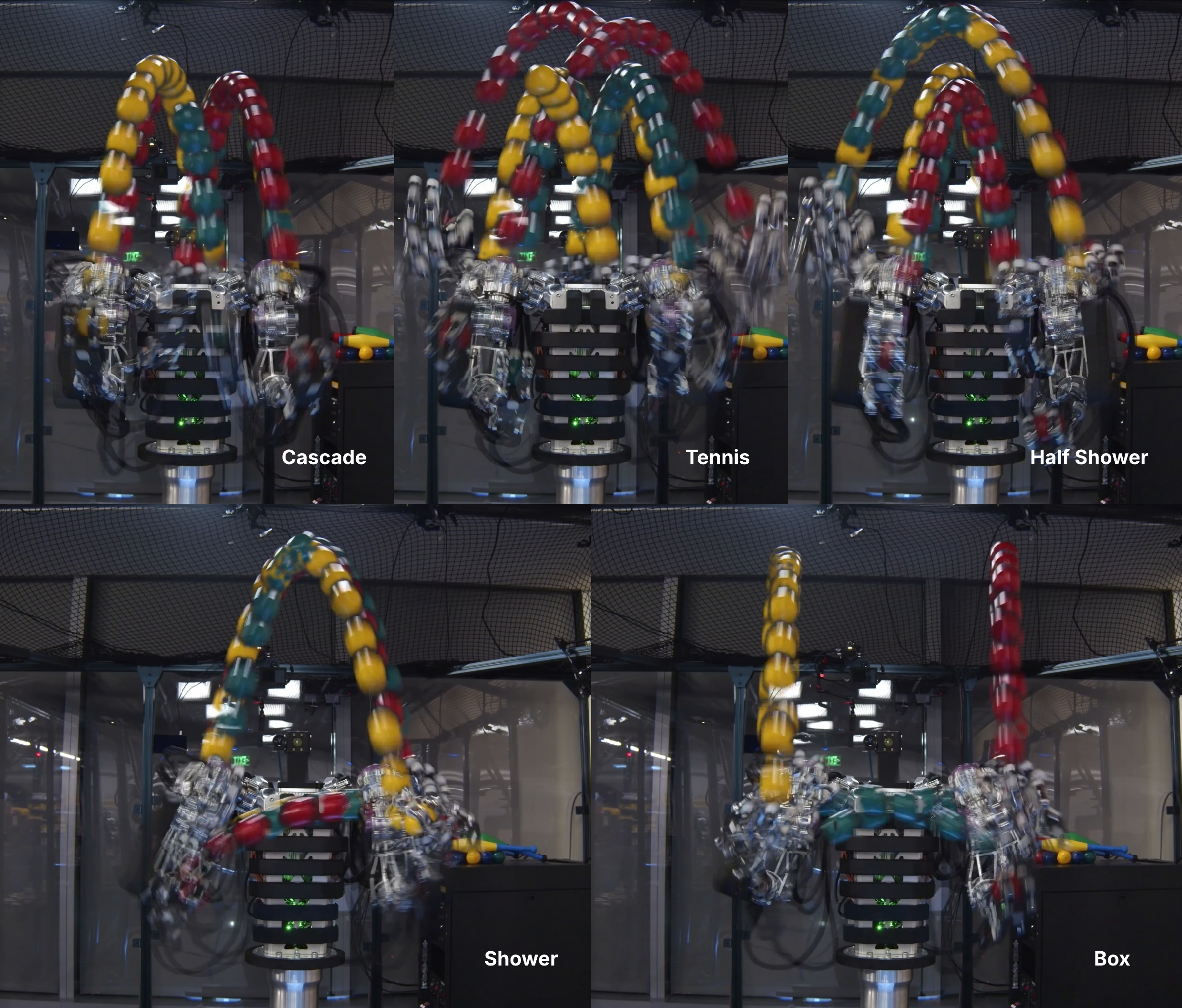}%
    }
    \caption{\textbf{Robot juggling five distinct three-ball patterns.} The bimanual robot AthenaZero continuously juggles the cascade, half-shower, tennis, shower, and box patterns, leveraging onboard vision and multi-fingered hands.}
    \label{movie:juggle}
\end{movie*}
\begin{table*}[t]
\centering
\caption{\textbf{Comparison of representative robotic juggling approaches.}}
\label{tab:juggling_comparison}

\begin{tblr}{
    width=\linewidth,
    colspec={
        X[1.55,l]
        X[1.0,l]
        X[1.0,l]
        X[1.30,l]
        X[1.05,l]
        X[1.55,l]
    },
    cells={m},
    row{1}={font=\bfseries},
    row{2,4,6,8}={bg=gray!15},
    row{9}={font=\bfseries},
    column{1}={halign=l},
    column{2}={halign=l},
    column{4}={halign=l},
    column{5}={halign=l},
    column{7}={halign=l},
    rowsep=3pt,
    colsep=5pt,
}
\toprule

Work &
Interaction &
{Execution\\feedback} &
{On-robot\\learning} &
{Physical\\experience} &
Demonstration \\

\midrule

Buehler et al.~\cite{buehler1994planning} &
Paddle &
Closed-loop &
-- &
-- &
1-ball bounce juggling \\

Rizzi and Koditschek~\cite{rizzi1993further} &
Paddle &
Closed-loop &
-- &
-- &
2-ball bounce juggling \\

Aboaf et al.~\cite{aboaf1989task} &
Paddle &
Closed-loop &
Parametric &
160 attempts &
1-ball bounce juggling \\

Schaal and Atkeson~\cite{schaal1994robot} &
Stick &
Closed-loop &
Real-time,
memory-based &
40--100 attempts &
Devil sticking \\

Ploeger et al.~\cite{ploeger2021high} &
Cup &
Open-loop &
Episodic, reinforcement learning &
500 attempts &
2-ball toss juggling \\

Ploeger and Peters~\cite{ploeger2026task} &
Cup &
Open-loop &
{Episodic, regularized\\memory-based} &
1--2 attempts &
5-ball cascade \\

Kizaki and Namiki~\cite{juzaju2012two} &
{Multi-fingered\\hands} &
Closed-loop &
-- &
-- &
2-ball cascade \\

Ours &
{Multi-fingered\\hands} &
Closed-loop &
{Real-time, regularized\\memory-based} &
7--10 attempts &
{3-ball cascade, \\half shower, tennis, shower, box } \\

\bottomrule
\end{tblr}
\end{table*}

We show that a robot system can be engineered to overcome these discrepancies through learning from its own physical experience. Our bimanual robot learns to juggle five distinct three-ball patterns---cascade, tennis, half-shower, shower, and box---with less than five minutes of on-robot learning. Importantly, this level of performance did not require engineering an accurate prior model. The prior model used in this study could not complete even a single cycle of any juggling pattern. Instead, we designed the system to continually improve its behavior from physical experience while building upon the knowledge already provided by this imperfect model.

We found rapid on-robot learning especially useful since our physical system (and most robots) do not remain exactly the same over time. We observed this directly in our juggling experiments. A behavior that performed robustly on one day could consistently fail on another, even though no intentional change had been made to the robot or balls. Small changes in the robot, objects, and their interactions were enough to affect performance. A system that can learn from its ongoing physical experience can respond to these changes as they arise, rather than requiring them to be anticipated during its initial design.  

On-robot learning, however, must make efficient use of physical experience. Every interaction takes real time, making extensive exploration prohibitively slow, and every attempted behavior is executed by the physical system, making unsafe exploration unacceptable. Starting from scratch is therefore neither practical nor necessary. Although our prior model could not produce successful juggling, we found that it still provided extremely useful information for learning. This suggests that the usefulness of a prior should not be judged solely by its zero-shot task performance. Rather than treating the prior merely as a starting point to be
replaced as experience accumulates, the more important question is how new experience can most effectively build upon the knowledge the robot already has.

We address this question through \emph{regularized memory-based learning}. The robot accumulates memories of its physical experience while retaining its prior model. Where relevant experience is available, these memories are used to construct a local model that reflects what has actually occurred on the robot. Where experience is sparse, the prior continues to provide guidance. New experience can therefore correct the prior where necessary without discarding useful knowledge elsewhere, enabling the robot to learn efficiently from each interaction without resorting to uninformed exploration over the full space of possible behaviors.

Efficient learning alone, however, is not sufficient for continual on-robot learning. Safety becomes particularly challenging when learning high-performance behaviors. Conservative restrictions on workspace, speed, or actuation can make on-robot learning safe, but can also exclude the dynamic motions required for the task. Learning high-performance dynamic behaviors therefore requires knowing how far the robot can push its motion while retaining the ability to safely execute what comes next.

Juggling makes this requirement particularly clear. A throw that is feasible by itself can leave an arm in a state from which the subsequent catch would require violating joint or actuator limits. We address this problem by constructing a \emph{mutually reachable set}, a precomputed constraint that ensures successive throws and catches can be safely composed without driving either arm into a state from which its next action would require violating its physical limits. This allows the robot to repeatedly practice high-performance behaviors without relying on overly conservative motion restrictions.

Together, these ideas enable a robot to continually improve from limited physical experience while safely learning high-performance behaviors. 

\subsection{Related Work}
Robotic juggling has long served as a testbed for dynamic manipulation~\cite{mason1993dynamic,beek1995science,schaal1994robot,buehler1994planning,burridge1999sequential}. Early systems demonstrated remarkably dynamic and robust juggling without on-robot learning, and in some cases without feedback during execution. A pioneering example is Shannon's entirely open-loop juggling machine, which achieved repeatable behavior through carefully designed mechanics that made the interaction between the robot and object passively stable~\cite{beek1995science}. This principle was later explored in robotic systems capable of a broader range of open-loop juggling behaviors~\cite{schaal1993open}. Other systems used visual feedback to continuously correct the robot's motion based on observed ball states, designing feedback laws that stabilize the repeated interaction~\cite{buehler1994planning,rizzi1993further}. 

Across these approaches, specialized end-effectors such as paddles and cups have played an important role in making ball contact predictable and reducing sensitivity to impact uncertainty. Such mechanical design can provide robustness without requiring physical learning. Multi-fingered hands provide a more general interaction interface, but expose the behavior to contact effects arising from friction, compliance, impact, and release timing that are difficult to model accurately \cite{juzaju2012two}. Together, prior works demonstrate that highly dynamic juggling can be achieved robustly without on-robot learning when the interaction between the robot and object is sufficiently predictable. Our work instead uses physical experience to overcome the remaining mismatch, enabling the same multi-fingered hands to learn and execute a diverse set of juggling patterns involving different throwing and catching motions and contact geometries.

On-robot learning has also been part of robotic juggling from its early development. Aboaf et al.~\cite{aboaf1989task} used iterative learning at the task level to improve open-loop throwing behavior, and also investigated learning parametric models of the system to improve subsequent feedback control. Schaal and Atkeson~\cite{atkeson1997locallycontrol,schaal1994robot} introduced memory-based, nonparametric approaches that retain and reuse physical experience to improve closed-loop behavior, providing the foundation for the learning approach developed in this work. More recently, on-robot reinforcement learning has been explored for robotic juggling~\cite{ploeger2021high}.

Particularly relevant to our work is the recent method of Ploeger and Peters~\cite{ploeger2026task}, which was developed independently and shares a closely related learning principle. Like our method, it regularizes a local model learned from physical experience using first-order information from an analytical prior, providing guidance for how to modify the command when nearby physical experience is sparse. The two approaches differ primarily in how this learning is incorporated into the overall juggling system. Ploeger and Peters associate corrections with individual throws in an open-loop sequence, whereas our learner operates on reusable skills whose commands are conditioned on the observed state resulting from preceding actions. Experience associated with a skill can therefore be reused under different incoming conditions and across different juggling patterns. This closed-loop formulation also allows learning to continue online during juggling, whereas Ploeger and Peters update the open-loop sequence episodically between trials. 
\begin{figure*}[t]
    \centering
    \includegraphics[width=\linewidth]{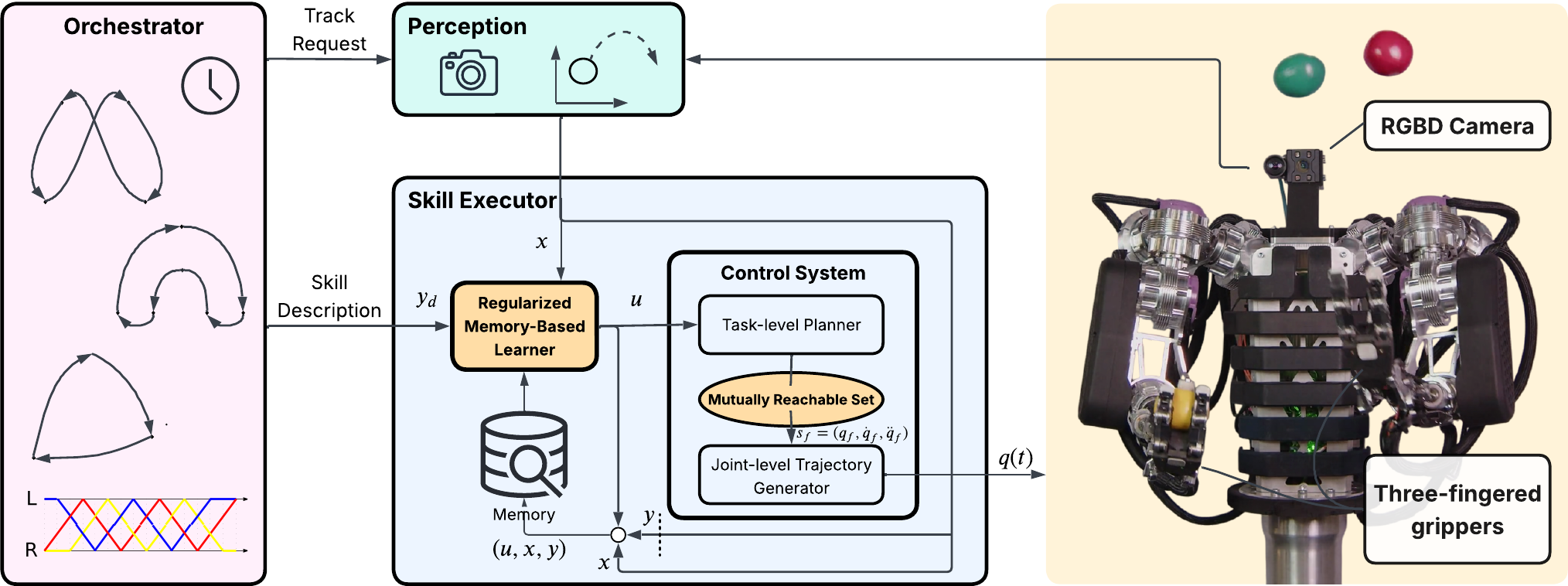}
    \caption{\textbf{Hierarchical, skill-based system architecture and real-time information flow for on-robot learning.} The \textit{Orchestrator} dispatches skill descriptions to the \textit{Skill Executor}, where the target outcome ($y_d$) is decoded. The \textit{Regularized Memory-Based Learner} uses the experience \textit{Memory} to adapt the command ($u$), while the model-based \textit{Control System} generates joint trajectories for the physical robot. A precomputed \textit{Mutually Reachable Set} constrains the task-level plan so that successive skills can be safely composed without violating the robot's physical limits. An asynchronous \textit{Perception} module tracks object states ($x$) to provide feedback for closed-loop execution. The resulting outcome ($y$) is stored in memory and immediately becomes available for refining commands.}
    \label{fig:system_overview}
\end{figure*}

Table~\ref{tab:juggling_comparison} summarizes the learning and execution characteristics of representative robotic juggling systems. Beyond juggling, we discuss broader connections to the on-robot/online learning paradigms in the Supplementary Materials.

\begin{table}[t] 
    \centering
    \rowcolors{2}{white}{gray!15}
    \caption{\textbf{Juggling skill description parameters}}
    \label{tab:skill_param}
    
    \begin{tabularx}{\columnwidth}{l l X}
        \toprule
        \textbf{Parameter} & \textbf{Type / Set} & \textbf{Description} \\
        \midrule
        Type            & $\in \{\texttt{throw},~\texttt{catch}\}$ & High-level robotic action \\
        Object ID       & $\in \{0,~1,~2\}$                       & Targeted juggling ball index \\
        Task Identifier & String                                  & Semantic routing label (e.g., \texttt{toR3fromL1}) \\
        Duration        & Float (seconds)                         & Target execution timeframe (e.g., 0.35) \\
        \bottomrule
    \end{tabularx}
\end{table}

\section{Results}
\label{sec:results}


Movie~\ref{movie:juggle} demonstrates the performance achieved by the current system and our learning framework. The robot autonomously juggles three balls using two multi-fingered hands and head-mounted vision, performing five distinct patterns: cascade, half-shower, tennis, shower, and box, as well as transitions between patterns. The movie provides a visual overview of the distinct throwing and catching motions that characterize each pattern and transition.

These behaviors are learned within minutes of physical interaction. Excluding time spent on manual resets, during which a human retrieves dropped balls and returns them to the robot, the robot learns cascade juggling in 53s of physical interaction on average and the continuous tennis--half-shower--cascade sequence in 75s on average. For the more challenging shower and box patterns, performance reaches its observed plateau after approximately 30s and 60s of physical interaction, respectively. Readers are encouraged to refer to Supplementary Movies S1--S4 to see the learning process as these behaviors are acquired in real time.

We first present an overview of the system, followed by an evaluation of the on-robot learning process and the performance achieved across the different juggling patterns.

\subsection{System Overview}
\label{sec:sys_overview}
The bimanual robot platform used in this work is AthenaZero, a custom-built, low-inertia manipulator developed at the RAI Institute \cite{athenazero2024blog}. 
Each hand features a three-fingered gripper; each underactuated, two-degree-of-freedom finger is driven by a remote actuator via a Bowden cable transmission, enabling a full opening and closing cycle each in approximately 80 ms.
The juggling objects are commercially available HB Juggling Balls (Higgins Brothers, model \#1003), each weighing 130~g and measuring 63.5~mm in diameter. The balls are soft, deformable beanbags that dissipate substantial energy upon impact and exhibit little rebound during contact with the hands.
The sensor suite used for ball tracking, rigidly mounted at the top of the torso, comprises a Lucid Helios2 Wide Time-of-Flight camera for depth mapping and a Lucid Triton RGB camera for object segmentation. The cameras are synchronized to operate at 30 Hz. The total latency, measured from the physical scene capture to the computed 3-DoF tracked position of the juggling balls, is approximately 0.1 s. The tracking system can be configured for different objects using a few point annotations in several example images during initial setup; the full pipeline is described in Section~4\ref{sec:perception}. The overall control system, illustrated in Figure~\ref{fig:system_overview}, relies on a hierarchical, skill-based architecture implemented within ROS2 Humble~\cite{mac2022ros2}. To support asynchronous data flows and multi-rate execution, each core block is decoupled into an independent ROS node. 

At the highest level, an \textbf{Orchestrator} asynchronously initiates a predefined sequence of juggling skills to the left and right arms according to a global wall-clock schedule. This timing maintains the temporal phase of the juggling pattern. If a skill takes longer than its scheduled duration, subsequent skills are not delayed by the same amount. Instead, the orchestrator maintains the original schedule, requiring subsequent motions to compensate for the accumulated timing error.

The orchestrator also monitors whether a juggling attempt can continue. An attempt is terminated when either the perception module cannot detect the ball required for the next catch or the Skill Executor determines that no feasible solution exists for the requested command. In practice, both cases often originate from an inaccurate preceding throw. A ball thrown too far may lie outside the reachable workspace, leaving no feasible solution for sustaining the juggling sequence. Even when the ball remains reachable, the desired catch may not be achievable at exactly the scheduled time under the robot's kinodynamic limits. In this case, the low-level trajectory generator still executes a feasible motion that reaches the target as close to the desired timing as possible. The resulting timing error may cause the robot to miss the catch, in which case perception subsequently reports the missing ball as a failure. Alternatively, the ball may still be caught but poorly aligned in the hand. In this case, juggling continues, but the poor alignment may affect the next throw.

The failures above are allowed to occur. What cannot be allowed is a motion that risks physical damage to the robot. For example, a throw may itself satisfy all the required kinodynamic limits, yet leave some joints moving toward its position limit at a velocity from which it cannot safely decelerate before running into the joint limit. We refer to preventing such situations as \emph{safety}. To prevent the robot from entering such states, the control system explicitly incorporates a precomputed \textbf{Mutually Reachable Set} as a strict constraint, detailed in Section~4\ref{sec:safe_action_space}. This allows the robot to learn and execute highly dynamic motions close to its physical limits without relying on overly conservative restrictions on speed.

Each skill initiated by the Orchestrator is defined by four parameters, as summarized in Table~\ref{tab:skill_param} and further detailed in the Supplementary Material. Based on these parameters, the execution pipeline processes the skill according to its designated type. For a \texttt{throw} skill, the \textbf{Regularized Memory-Based Learner} receives the current state $\mathbf{x}$, decodes the desired outcome $\mathbf{y}_d$ from the task identifier, and references the experience memory $\mathcal{D} = \{(\mathbf{x}_i, \mathbf{u}_i, \mathbf{y}_i)\}_{i=1}^{N}$ to output a corrected task-level command $\mathbf{u}$ for the current throw. Detailed definitions of $\mathbf{x}$, $\mathbf{u}$, and
$\mathbf{y}$ are provided in Section~4\ref{sec:memory_learning}.
Briefly, $\mathbf{x}$ represents the state of the ball at the beginning
of the current throw, $\mathbf{u}$ is the commanded landing position,
and $\mathbf{y}$ is the observed landing position. The \textbf{Task-Level Planner} then translates the command $\mathbf{u}$ into target joint-space positions, velocities, and accelerations  while enforcing the mutually reachable set constraint. Finally, the \textbf{Joint-Level Trajectory Generator} computes a trajectory toward these target values, which is then evaluated at 1 kHz to provide references to a standard inverse-dynamics-based tracking controller. This entire inference and planning loop is executed once at the onset of every throw. For a \texttt{catch} skill, however, the learning module is bypassed. Instead, the system continuously replans the target joint state and trajectory using the latest ball-state estimate from the perception module, until 0.1s before the scheduled catch time. This cutoff balances the improved estimation accuracy provided by additional observations against the reduced time available for the arm to reach the updated catch state.

Notably, the perception module tracks individual balls asynchronously with respect to the control system. This asynchronous execution is useful because the flight phases overlap: an object enters free flight before the receiving hand finishes its ongoing throw and transitions to a catch. Once the orchestrator receives a completion callback from a \texttt{throw} skill, it issues a tracking request to the perception module for the specific ball ID associated with that throw.

\subsection{On-Robot Learning of Juggling Skills}
\begin{figure*}[h!]
    \centering
    \includegraphics[width=\linewidth]{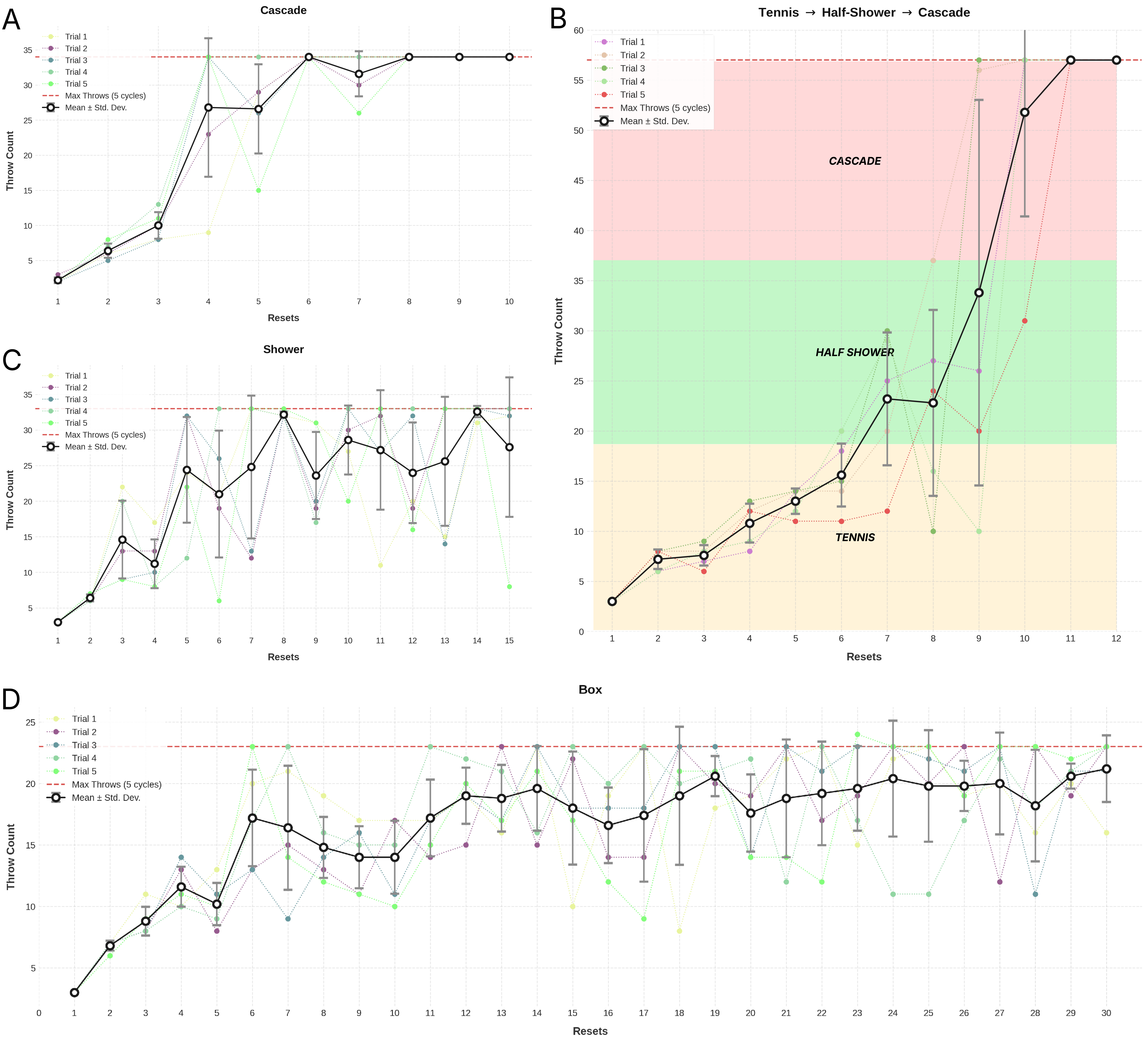}
    \caption{\textbf{Learning curves across distinct juggling patterns.}
The x-axis tracks successive juggling runs separated by manual resets, and the y-axis denotes the consecutive throw count attained in each run. Learning curves of individual trials (1--5), where each trial starts learning from scratch and proceeds across a sequence of resets, are shown in lighter colors, with the solid black line indicating the mean $\pm$ standard deviation. The dashed red line marks the consecutive throw count corresponding to five complete cycles of each juggling pattern. (A) The cascade is rapidly acquired, consistently reaching the five-cycle target after several resets. (B) During continuous learning across Tennis $\rightarrow$ Half-Shower $\rightarrow$ Cascade, experience is retained as the robot transitions between patterns, with the cascade acquired rapidly after learning tennis and half-shower. (C,D) Performance on the more difficult shower and box patterns progressively improves with physical experience, although neither pattern consistently reaches the five-cycle target.}
    \label{fig:learning_curve_all}
\end{figure*}

Our primary contribution is enabling a robot to rapidly learn juggling skills with minimal real-world experience. We first demonstrate this capability using the most fundamental juggling pattern: the three-ball cascade. As shown in Figure \ref{fig:learning_curve_all}A, the robot learns to sustain a cascade for five cycles after only seven resets (defined as a human manually placing the three balls back into the robot's hands following a failure). To evaluate consistency, we conducted five independent trials from scratch, with no memory retained between runs. Across all trials, the robot consistently learned to juggle the cascade pattern by the 8th attempt and successfully repeated the performance three consecutive times, requiring approximately 5 minutes of total wall-clock time, including manual resets.

\begin{figure*}[h!]
    \centering
    \includegraphics[width=\linewidth]{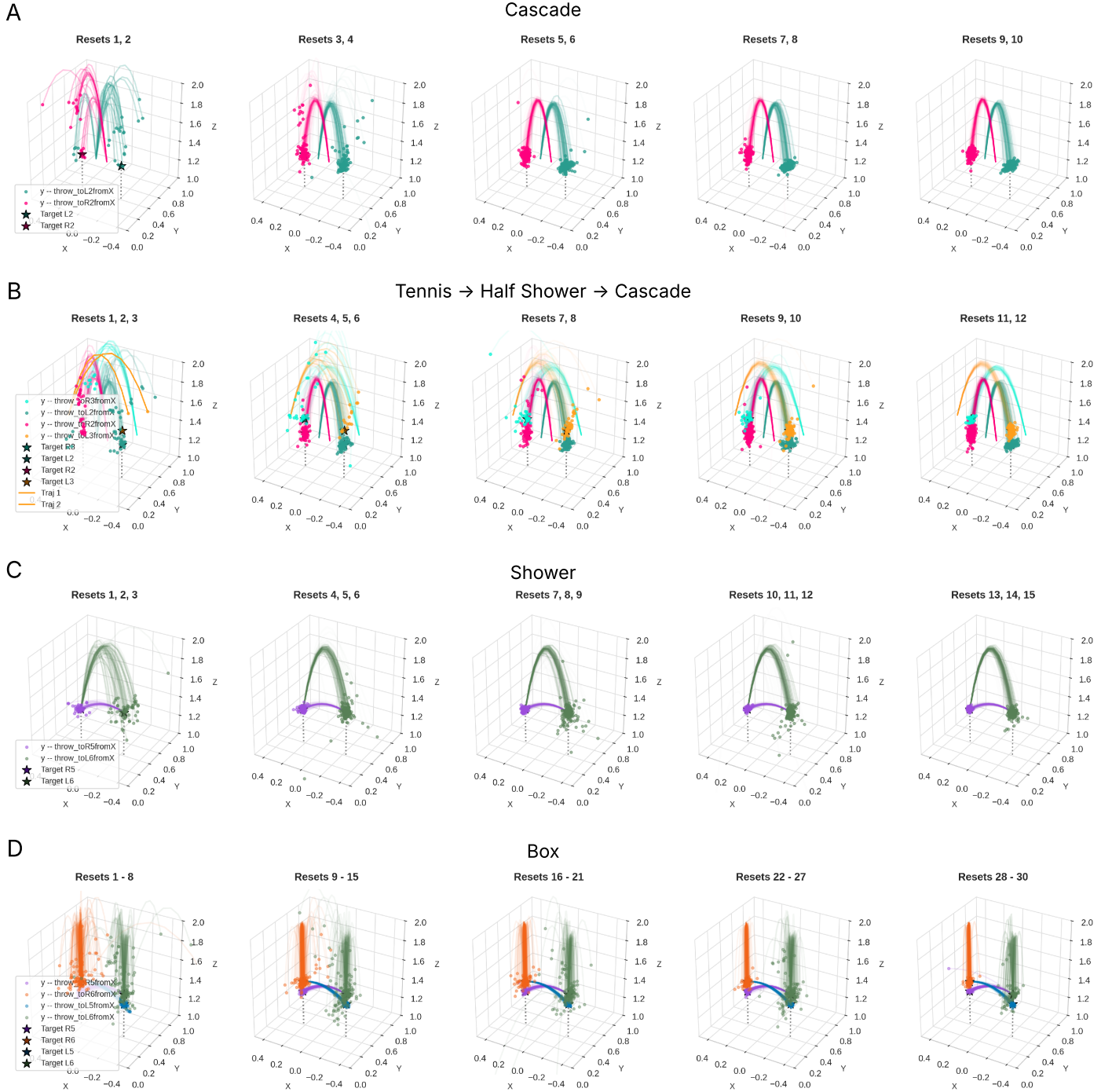}
    \caption{\textbf{Evolution of 3D ball trajectories and landing distributions during on-robot learning.} Each row tracks the spatial distribution of ballistic flight arcs, and ball landing locations over sequential resets for different juggling patterns: (\textbf{A}) Cascade, (\textbf{B}) a multi-pattern transition sequence (Tennis $\rightarrow$ Half-Shower $\rightarrow$ Cascade), (\textbf{C}) Shower, and (\textbf{D}) Box. Moving from left to right across successive resets, the scatter points become tightly clustered around their designated targets, with reduced variance in the flight trajectories, demonstrating the sample-efficient improvement achieved through real-time adaptation.}
    \label{fig:ball_distribution_all}
\end{figure*}

The ball landing distributions in Fig.~\ref{fig:ball_distribution_all}A clearly show how throwing accuracy improves with physical experience. At the beginning of learning, when little real-world experience is available and the behavior is guided primarily by the prior model, the landing distribution is far from the desired target. As experience accumulates, the robot rapidly corrects this mismatch and learns to produce accurate throws from different catch states. Importantly, learning occurs during ongoing juggling rather than only between attempts. After each throw, the resulting outcome is observed and the experience tuple $(x_i,u_i,y_i)$ is immediately appended to the memory buffer $\mathcal{D}$, allowing that experience to influence subsequent throws. Consequently, as the robot sustains juggling for longer, less time is spent on manual resets and physical experience accumulates more rapidly, creating a positive feedback loop that accelerates real-time learning.

Notably, the underlying prior model is sufficiently inaccurate that the robot cannot sustain even a single juggling cycle without adaptation. Despite this inaccuracy, the prior provides useful information about how changes in the task-level command are likely to affect the outcome, allowing the learner to begin improving the behavior even when no physical experience is initially available. In contrast, memory-based learning without such prior information generally requires an initial set of physical samples, often collected through dedicated or random exploration, before useful command updates can be estimated~\cite{schaal1994robot}. A distinct advantage of our approach is that learning can begin immediately without requiring such a dedicated initialization or exploration phase.

\begin{figure}[t]
\centering
\includegraphics[width=\linewidth]{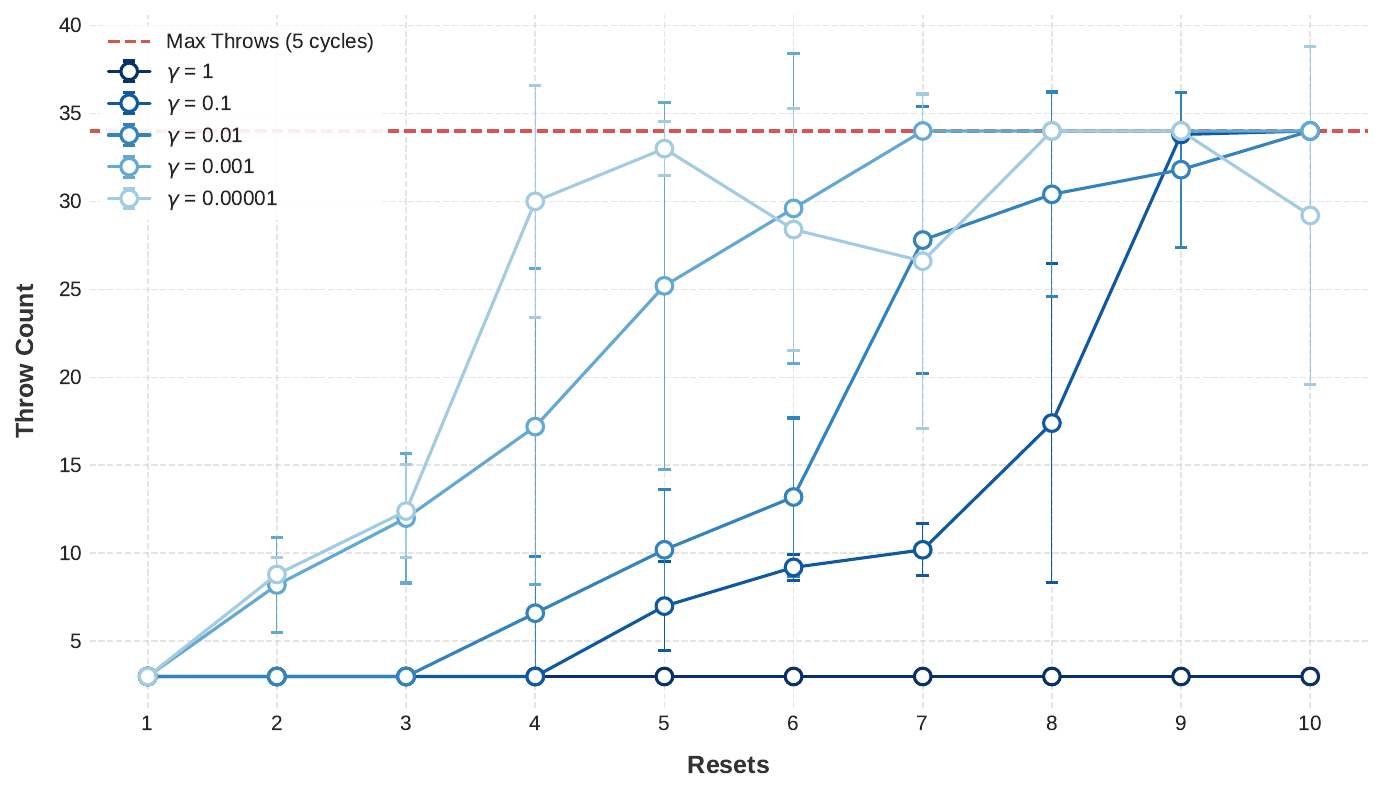}
\caption{\textbf{Impact of prior regularization on learning curves.} Evaluation is conducted on the cascade juggling pattern across varying choices of the prior regularization coefficient $\gamma >0$. A nonzero $\gamma$ regularizes the locally fitted model toward the prior, with larger values placing greater weight on the prior relative to the observed experience. For each coefficient value, the robot executed five independent trials, each learning from scratch. The curves show the mean $\pm$ standard deviation across trials to highlight performance trends.}
\label{fig:learning_curve_gamma}
\end{figure}

As detailed in Section~4\ref{sec:memory_learning}, the key to learning rapidly when little physical experience is initially available is how the learner combines information from the prior model with real-world experience stored in memory. The regularization factor $\gamma > 0$ controls this balance. A larger $\gamma$ causes the learner to rely more strongly on the prior, resulting in more stable but slower adaptation to physical experience, whereas a smaller $\gamma$ allows physical experience to have a stronger influence but also makes learning more sensitive to sparse or noisy observations. To evaluate this trade-off, we repeated the cascade learning experiment while varying $\gamma$. As shown in Figure~\ref{fig:learning_curve_gamma}, excessive regularization substantially slows learning, while insufficient regularization produces more variable and less consistent learning. For all experiments shown in Figure~\ref{fig:learning_curve_all}, we use $\gamma=0.001$, which provides a balance between these two effects.



\subsection{Reusability of Skills across Diverse Juggling Patterns}

A key component of our approach involves decomposing the continuous, long-horizon task of multi-ball juggling into a sequence of discrete, reusable skills. For instance, as detailed in the supplementary material, the throwing skills required for the tennis pattern encompass those constituting both the cascade and half-shower. Consequently, the experience acquired while learning the tennis pattern can be transferred to facilitate the learning of the cascade and half-shower. Figure \ref{fig:learning_curve_all}B illustrates the learning trajectory of the robot executing a sequential transition from tennis to half-shower and finally to cascade within a single continuous run, where each pattern is sustained for five cycles.

As shown in Figure \ref{fig:learning_curve_all}B, the robot successfully masters tennis within approximately 7–8 resets, consistent with the isolated cascade baseline. Upon transitioning to the subsequent patterns, the robot rapidly achieves proficiency in both the half-shower and cascade within only 2–3 additional runs. The robot does not immediately succeed at the subsequent patterns because the state ($\mathbf{x}$) distribution—specifically, the catching locations and ball landing positions as detailed in the supplementary material—varies across different patterns despite sharing the same underlying skills. Consequently, the learner handles this distribution shift by simply acquiring a small amount of additional experience on an on-demand basis to complete the full multi-pattern sequence. This rapid adaptation across shared skills underscores the benefit of learning a closed-loop stateful policy rather than an open-loop mapping. Without state feedback $\mathbf{x}$, an open-loop controller cannot differentiate the required control inputs $\mathbf{u}$ for identical skills executed across distinct states, which would otherwise mandate learning entirely separate, dedicated skills from scratch for each pattern.


\subsection{Learning with Extreme Precision}
The shower and box patterns demand significantly higher throwing precision than the other patterns we implemented. This requirement is dictated by a fundamental physical constraint: the short flight time of the side throws. Because the flight duration of the side throws is shorter than our robotic system's sensorimotor and visual processing latency, reliable real-time visual tracking during flight is impossible; the system cannot acquire enough sequential observations before the catch to generate a low-variance prediction of the ball's landing trajectory.
This constraint forces the catching phase for side-thrown balls to operate open-loop with respect to visual feedback of the ball, with the receiving hand catching at a fixed target position and at the desired timing for the other arm's throw. Although the visual feedback arrives too late to correct the catch itself, the landing position of the thrown ball can still be measured and stored in memory. The learner can use this information to improve the accuracy of subsequent side throws, progressively increasing the throwing precision required for the open-loop catch.

As illustrated in Figure \ref{fig:ball_distribution_all}, the ball landing distribution successfully converges to a highly concentrated target zone over successive resets, demonstrating the effectiveness of the adaptation. Nevertheless, as shown in the learning curves of Figures \ref{fig:learning_curve_all}C and \ref{fig:learning_curve_all}D, the shower and box patterns do not achieve the same level of deterministic, zero-failure convergence across all trials as the cascade, half-shower, or tennis. While the robot occasionally completes the full five-cycle sequence for both the shower and box, the overall success probability remains lower. This performance plateau suggests that the purely open-loop catching policy may limit the robustness of these more complex patterns. We further discuss this limitation and its implications in the Discussion section.

\subsection{Safety}

The experiments above require the task-level planner to continuously adjust throwing and catching motions in real time as the learned commands and observed ball states change. Although an individual target transition state may satisfy the instantaneous joint-state bounds, i.e., the position, velocity, and acceleration limits, it may not be reachable from the preceding state or may leave the arm unable to safely execute a subsequent motion without, for example, running into a joint limit. The Mutually Reachable Set (MRS) is imposed on the task-level planner to prevent such states from being commanded during on-robot learning.

To evaluate how often this additional constraint is relevant in practice, we performed a retrospective ablation over all target-state planner queries generated during the real-world juggling experiments presented above. For each recorded command from the learner, we recomputed the target joint transition state after removing the MRS constraint while retaining the instantaneous joint-state bounds. Thus, the planner without the MRS constraint was still required to satisfy the robot's position, velocity, and acceleration limits at the target state.
We then tested whether each resulting target state was reachable from the preceding transition state and viable for subsequent motion using the same joint-level trajectory generator employed during the experiments. The trajectory generator, Ruckig~\cite{berscheid2021jerk}, solves time-optimal trajectory generation subject to joint position, velocity, acceleration, and jerk limits using analytic solution profiles. We determine reachability and viability from whether Ruckig finds feasible trajectories to and from the target state, respectively; the solver reports an error when no such trajectory exists. We classify a solution as safe when it is both reachable and viable.

\begin{table*}[t]
\centering
\caption{
\textbf{Reachability and viability of task-level planner solutions with and without
the Mutually Reachable Set (MRS) constraint across all juggling experiments.}
The unconstrained solutions are obtained retrospectively from the same planner
queries after removing the MRS constraint. Solutions outside the MRS are
further classified as safe when they are both reachable from the preceding
transition state and viable for subsequent motion, and unsafe otherwise.
Because the MRS is a conservative inner approximation, states outside the set
are not necessarily unsafe. Percentages are computed with respect to the total
number of planner queries for each pattern.
}
\label{tab:mrs_ablation}

\begin{tblr}{
    width=\linewidth,
    colspec={
        X[2.5,l]
        X[1.00,c]
        X[1.35,c]
        X[1.00,c]
        X[1.00,c]
        X[1.00,c]
    },
    cells={m},
    row{1}={font=\bfseries},
    column{1}={halign=l},
    rowsep=3pt,
    colsep=5pt,
}
\toprule

Juggling pattern &
Cascade &
{Tennis\\
 $\rightarrow$ Half-Shower\\
 $\rightarrow$ Cascade} &
Shower &
Box &
All \\

\midrule

\shortstack[l]{
    {Number of experience (throw) data}\\
    {over all runs}
} &
1198 &
2256 &
1634 &
2490 &
7578 \\

\midrule

\SetCell[c=6]{l}
\shortstack[l]{
    \textit{Without MRS constraint}\\
    \textit{(retrospective ablation)}
}
& & & & & \\

Inside MRS &
16 (1.3\%) &
12 (0.5\%) &
25 (1.5\%) &
14 (0.6\%) &
67 (0.9\%) \\

\SetRow{bg=gray!15}
Outside MRS &
1182 (98.7\%) &
2244 (99.5\%) &
1609 (98.5\%) &
2476 (99.4\%) &
7511 (99.1\%) \\

\quad Safe &
5 (0.4\%) &
0 (0.0\%) &
761 (46.6\%) &
0 (0.0\%) &
766 (10.1\%) \\

\SetRow{bg=gray!15}
\quad \textbf{Unsafe} &
\textbf{1177 (98.2\%)} &
\textbf{2244 (99.5\%)} &
\textbf{848 (51.9\%)} &
\textbf{2476 (99.4\%)} &
\textbf{6745 (89.0\%)} \\

\qquad Reachable, not viable &
1102 (92.0\%) &
2117 (93.8\%) &
70 (4.3\%) &
141 (5.7\%) &
3430 (45.3\%) \\

\SetRow{bg=gray!15}
\qquad Viable, not reachable &
49 (4.1\%) &
38 (1.7\%) &
778 (47.6\%) &
2326 (93.4\%) &
3191 (42.1\%) \\

\qquad Neither reachable nor viable &
26 (2.2\%) &
89 (3.9\%) &
0 (0.0\%) &
9 (0.4\%) &
124 (1.6\%) \\

\midrule

\SetCell[c=6]{l}
\textit{With MRS constraint}
& & & & & \\

\SetRow{bg=gray!15}
\textbf{Inside MRS, safe} &
\textbf{1198 (100\%)} &
\textbf{2256 (100\%)} &
\textbf{1634 (100\%)} &
\textbf{2490 (100\%)} &
\textbf{7578 (100\%)} \\

\bottomrule
\end{tblr}
\end{table*}

Table~\ref{tab:mrs_ablation} shows that the MRS constraint is active in a
meaningful region of the motions encountered during juggling. Without the
constraint, 99.1\% of the task-level solutions lie outside the precomputed
MRS. Because the MRS is a strictly inner approximation of the true maximally mutually reachable set, leaving the set
does not always imply that a solution is unsafe. We therefore separately
evaluate reachability and viability as described above. Across all 7,578 planner queries, 89.0\%
of the unconstrained solutions are found to be unsafe, i.e., either unreachable from the preceding
transition state, non-viable for subsequent motion, or both. In contrast, all
solutions generated with the MRS constraint remain within the set and are
therefore both reachable and viable, safe by construction.

The dominant failure mode also varies substantially across juggling patterns. For the cascade and the continuous tennis--half-shower--cascade sequence, unconstrained solutions are predominantly reachable but not viable, indicating that the arm can reach the desired state but may subsequently be unable to move away without violating its limits. For the box pattern, the opposite failure mode dominates: unconstrained target states are generally viable but cannot be safely reached from the preceding state.
Interestingly, except for the shower pattern, membership in the MRS closely matched the transition-specific safety test. Among unconstrained solutions outside the MRS, 99.6\% were unsafe for cascade, and 100\% were unsafe for both the continuous tennis--half-shower--cascade sequence and box. The shower pattern was notably different: 761 of 1,609 solutions outside the MRS (47.3\%) remained reachable from the preceding state and viable for subsequent motion. This is consistent with the MRS being an inner approximation: states outside the set may still be safe for particular transitions, although for most behaviors considered here the MRS boundary closely separated safe and unsafe solutions. 

\begin{figure*}[t]
    \centering
    \includegraphics[width=\linewidth]{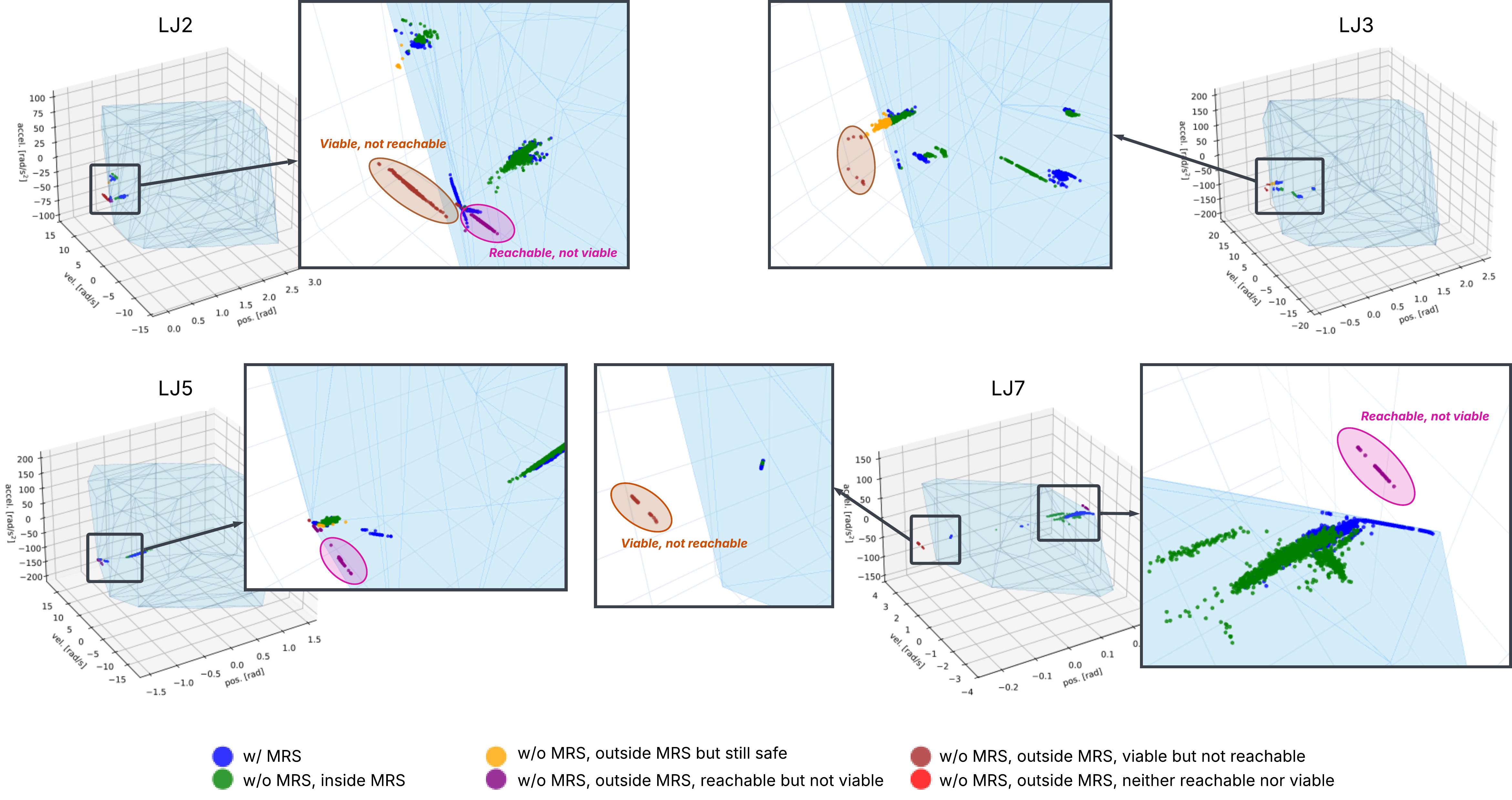}
    \caption{
    \textbf{Effect of the Mutually Reachable Set on target transition states during juggling.}
    The translucent blue polytopes show the precomputed MRS for representative
    left-arm joints in joint position, velocity, and acceleration space. Blue
    points denote target transition states generated with the MRS constraint.
    For the corresponding retrospective ablation without the constraint,
    green points remain inside the MRS, while states outside the set are
    classified as reachable and viable (orange), reachable but not viable
    (purple), viable but not reachable (brown), or neither reachable nor
    viable (red). Insets highlight regions in which the unconstrained
    task-level planner produces states outside the MRS. The examples show both
    potential major failure modes (without MRS) during juggling behaviors on AthenaZero: states that are reachable
    but not viable and states that are viable but not reachable.
    }
    \label{fig:safety_mrs}
\end{figure*}

Figure~\ref{fig:safety_mrs} visualizes representative examples of these failure modes in the joint transition-state space. The MRS-constrained solutions remain inside the precomputed polytopes, with many lying close to their boundaries, indicating that the MRS constraints are frequently active during juggling. In contrast, removing the MRS constraint allows the task-level planner to select target states beyond these boundaries, including both reachable-but-non-viable and viable-but-unreachable states.

We note that the figure may appear to contain a larger fraction of unconstrained solutions inside the MRS (green points) than reported in Table~\ref{tab:mrs_ablation}. This is because the figure shows the MRS and planner solutions joint-wise, whereas the table classifies the complete multi-joint target transition state. A target state is considered outside the MRS if the constraint is violated for any joint. Likewise, the full target state is considered reachable and viable only if these conditions hold for all joints; for example, if one joint is non-viable while another is unreachable, the complete target state is classified as neither reachable nor viable.
\section{Discussion}

Our experimental results across diverse juggling configurations demonstrate that a robot can rapidly acquire highly dynamic, long-horizon manipulation skills through physical interaction. Rather than learning these behaviors from scratch, our framework uses an imperfect prior to guide learning where physical experience is limited and increasingly relies on locally accumulated experience as it becomes available. Decomposing multi-ball juggling into reusable skills further allows experience acquired while learning one pattern to be reused when the same skills are encountered under different conditions or in other patterns. Together, these results demonstrate how prior knowledge and targeted physical experience can be combined to make on-robot learning practical for dynamic manipulation.


The choice of hardware also plays an important role in how much adaptation is required. Prior juggling systems have demonstrated remarkable robustness using specialized end-effectors, such as paddles and cups, that make object interaction more predictable. Such mechanical design can reduce sensitivity to contact uncertainty and, when the task permits it, may reduce or even eliminate the need for physical learning. Our use of soft, deformable balls similarly reduces rebound and dissipates energy upon impact, simplifying some aspects of catching. However, it does not eliminate contact uncertainty with multi-fingered hands. In particular, the deformable ball can undergo multiple intermittent contacts with the fingers during release, making the resulting throw sensitive to friction, compliance, and release timing. Harder or more elastic balls would alter these interaction dynamics, introducing greater rebound and potentially increasing sensitivity to impact conditions. Investigating whether the current learning framework can adapt to these more significant changes in object properties and the resulting model mismatch is an interesting direction for future work. Multi-fingered hands provide a more general interaction interface for executing different juggling patterns, but at the cost of more complex contact interactions. Our results suggest that mechanical design and on-robot adaptation need not be viewed as alternatives. Better mechanical design can make behavior more predictable, while on-robot adaptation can compensate for the remaining mismatch that is difficult to eliminate by design alone.

Beyond initial acquisition, we also observed throughout the development of the system that changes in physical conditions could affect previously learned behavior. Seemingly minor changes, such as the accumulation of dust or variations in humidity, appeared to alter the interaction between the hands and balls. In one particularly noticeable instance, cleaning the robot's palm and finger pads together with the juggling balls changed the contact conditions enough that previously learned behavior no longer performed reliably, and the robot had to adapt again from new physical experience. While these changes were not systematically evaluated in this work, they motivate the ability to continue adapting as an important consideration for maintaining high-performance manipulation over extended periods of real-world operation.

The remaining failures in the \textit{shower} and \textit{box} patterns highlight a limitation imposed by feedback latency. As discussed in the Results, the short flight times of the side throws prevent visual feedback from correcting the catch in time. Overcoming this limitation may require feedback at multiple levels and timescales. The task-level learner can improve throwing parameters over repeated interactions, while high-bandwidth proprioceptive or tactile feedback at the finger level could provide local reactive corrections during the catch. Combined with compliant finger actuation, such feedback could accommodate residual errors in ball position and timing, effectively widening the mechanical margin for error and reducing the reliance on precise open-loop catching. Separately, for the throw, tactile or proprioceptive sensing could provide a more direct estimate of the grasp state than vision. The input state to the learner could be augmented with this information, allowing the learner to account for grasp variations when determining the subsequent throw command.

Another direction is to design better low-level strategies and parameterizations not only for nominal task performance, but also to make their outcomes more repeatable and easier to improve through learning. The \textit{box} pattern provides an illustrative example. Learning the left-hand upper throw was considerably slower and more variable than learning the corresponding right-hand throw, despite the two skills being nominally similar (Figure~\ref{fig:learning_curve_all}D). We frequently observed the ball recontacting the fingers during release before clearing the hand, often introducing abrupt changes in the resulting throw. Such interactions can make the relationship between task-level commands and outcomes highly nonlinear or even discontinuous, making local learning considerably more difficult. A better low-level strategy and parameterization could avoid these sensitive interaction regimes and produce more consistent outcomes, making subsequent adaptation faster and easier. This suggests that the design of low-level behavior can itself play an important role in how effectively the robot can adapt from physical experience at the task level.

Moving forward, we aim to extend this framework to more dynamic and dexterous manipulation behaviors, such as club juggling. Unlike the ball-juggling behaviors considered here, club juggling requires controlling both position and orientation through coordinated interactions with the fingers and thumb, making an effective low-level strategy considerably more difficult to design by hand or through simple trajectory optimization. One possible approach is to replace the hand-designed task planner and trajectory generator with a goal-conditioned reinforcement learning (RL) policy that learns these low-level behaviors. The same high-level memory-based learner could then adapt the goal commands provided to this policy, using physical experience to rapidly improve performance in the real world.

Beyond the skill-level adaptation considered in this work, another direction is to extend our framework to tasks that require continuous adaptation or inference across sequences of multiple skills. Juggling is particularly amenable to our formulation because it provides natural boundaries between individual skills with relatively straightforward subgoals, such as where to throw or catch, and improving these individual skills strongly aligns with the global objective of sustaining the juggling pattern. In other manipulation tasks, a useful skill decomposition may be more difficult to define, and the appropriate subgoals for individual skills may not be apparent. For example, assembly may require adapting a sequence of approach, contact, alignment, and insertion behaviors while accounting for how the outcome of each affects those that follow. Forceful interaction may require continuous adaptation at a finer timescale, rather than updating the learned behavior only after the completion of a skill. An important open question is how to retain the computational efficiency and flexibility of memory-based learning for longer-horizon model-based inference or continuous real-time adaptation.

Safety also becomes more challenging when the task involves heavier objects or higher interaction forces. The mutually reachable set used here considers the robot's kinodynamic limits, while the interaction forces associated with the lightweight juggling balls do not materially affect safety in the present system. Extending this approach to more forceful interactions would require safety constraints that also account for the physical consequences of contact with objects and the environment. 

Extending the framework in these directions could broaden safe, sample-efficient learning and adaptation to a wider range of dynamic and contact-rich manipulation tasks, toward high-performance robot systems that continue to improve through physical experience.


\section{Materials and Methods}
\label{sec:methods}

This section details the core algorithmic framework enabling rapid, safe adaptation of juggling skills on the robot. Our framework is structured into three interconnected modules that operate with distinct timing, update rates, and levels of abstraction. First, Section~4\ref{sec:safe_action_space} develops a parameterized behavior generator for dynamically composing throws and catches on physical hardware. A high-level optimization-based task planner selects target joint states of each behavior, while a low-level trajectory generator realizes the corresponding high-speed robot motions. The MRS, the mutually reachable set denoted as $\mathcal{M}_S$, constrains the system to maintain kinodynamic feasibility across successive behaviors. 
Second, Section~4\ref{sec:memory_learning} details the high-level learning layer, introducing a regularized local linear regression framework that blends sparse memory tuples with first-order derivatives of a prior model to allow stable and rapid adaptation updates. Finally, Section~4\ref{sec:perception} outlines our object-agnostic vision engine, which combines prompt-driven instance segmentation with fast geometric registration to predict ballistic flight trajectories in real time.

\subsection{Safe Dynamic Composition of Task-Parameterized Trajectories}
\label{sec:safe_action_space}

As described in Section~2\ref{sec:sys_overview}, juggling is executed as a sequence of throwing and catching skills based on commands $\mathbf{u}$ from the learner. Each skill generates a joint-level trajectory connecting two transition ``joint'' states, defined here as commanded joint reference states consisting of joint position, velocity, and acceleration. These states are references sent to the trajectory tracking controller, rather than measured robot states or the task-level feedback state $\mathbf{x}$ used elsewhere in the paper. We represent the resulting sequence as
\begin{equation}
\mathbf{q}(\cdot \mid \mathbf{s}_{k-1}, \mathbf{s}_k(\mathbf{u}_k)) :
[0,t_f^k] \rightarrow
\mathcal{Q} \subseteq \mathbb{R}^n,
\qquad k=1,2,\ldots,
\end{equation}
where $\mathbf{u}_k$ is the task-level command for the $k$th skill,
$t_f^k$ is its duration, and $\mathcal{Q}$ denotes the configuration
space of the $n$-degree-of-freedom robot. The transition state is defined as
\begin{equation}
    \mathbf{s}
    =
    (\mathbf{q},\dot{\mathbf{q}},\ddot{\mathbf{q}})
    \in \mathcal{S}\subseteq\mathbb{R}^{3n},
\end{equation}
such that consecutive trajectories match in position ($\mathbf{q}$), velocity($\dot{\mathbf{q}}$), and
acceleration ($\ddot{\mathbf{q}}$) at the transition. Including acceleration in the transition state accounts for higher-order actuator dynamics, while requiring its continuity reflects the limited actuator bandwidth as described in the Supplementary Material.

Unlike a fixed open-loop sequence
$\mathbf{s}_0,\mathbf{s}_1,\mathbf{s}_2,\ldots$, the target transition
state $\mathbf{s}_k(\mathbf{u}_k)$ and the trajectory used to reach it
are replanned online based on the command
$\mathbf{u}_k$. As the learner updates these commands from physical
experience, the future transition state also changes
online. This flexibility enables the robot to continually adjust its
juggling behavior, but also requires it to safely compose trajectories
whose transition states are not known in advance. As discussed in
Section~2\ref{sec:sys_overview}, safe composition requires more than satisfying the robot's
physical limits along each individual trajectory: each transition state
must also allow subsequent trajectories to be executed without violating
those limits. The remainder of this section describes how we enforce
this requirement efficiently during online replanning.

Given an initial transition state
$\mathbf{s}_i=(\mathbf{q}_i,\dot{\mathbf{q}}_i,\ddot{\mathbf{q}}_i)$,
a target transition state
$\mathbf{s}_f=(\mathbf{q}_f,\dot{\mathbf{q}}_f,\ddot{\mathbf{q}}_f)$,
and a scheduled duration $T$ specified by the orchestrator, we connect
the two states using a minimum-duration jerk-constrained trajectory:
\begin{align}
\mathbf{q}^*(\cdot \mid \mathbf{s}_i,\mathbf{s}_f)
= \mathop{\arg\min}_{\mathbf{q}(t),~t_f} & \quad t_f \nonumber\\
\mathrm{s.t.} & \quad t_f \geq T, \nonumber\\
& ~(\mathbf{q}(0),\dot{\mathbf{q}}(0),\ddot{\mathbf{q}}(0))
    = \mathbf{s}_i, \label{eq:bc1}\\
& ~(\mathbf{q}(t_f),\dot{\mathbf{q}}(t_f),\ddot{\mathbf{q}}(t_f))
    = \mathbf{s}_f, \label{eq:bc2}\\
& ~\mathbf{q}_{\mathrm{min}}
    \leq \mathbf{q}(t) \leq \mathbf{q}_{\mathrm{max}},\\
& ~|\dot{\mathbf{q}}(t)| \leq \mathbf{v}_{\mathrm{max}},
    \quad
    |\ddot{\mathbf{q}}(t)| \leq \mathbf{a}_{\mathrm{max}},
    \label{eq:cons1}\\
& ~|\dddot{\mathbf{q}}(t)| \leq \mathbf{j}_{\mathrm{max}},
    \quad \forall t\in[0,t_f].
    \label{eq:cons2}
\end{align}
The bounds
$\mathbf{q}_{\mathrm{min}}$, $\mathbf{q}_{\mathrm{max}}$,
$\mathbf{v}_{\mathrm{max}}$, $\mathbf{a}_{\mathrm{max}}$, and
$\mathbf{j}_{\mathrm{max}}$ reflect the physical limits of the robot,
as detailed in the Supplementary Materials. We solve this trajectory
generation problem using Ruckig~\cite{berscheid2021jerk}.
The constraint $t_f \geq T$ preserves the global schedule whenever the target state can be reached at the scheduled time. If this is not possible under the trajectory constraints, the optimizer instead returns the shortest feasible duration $t_f > T$, introducing the timing error described in Section~2\ref{sec:sys_overview}.

We now characterize a set of transition states within which arbitrary
successive transitions can be safely composed under the robot's physical
limits.

\begin{figure}[t]
    \centering
    \includegraphics[width=\linewidth]{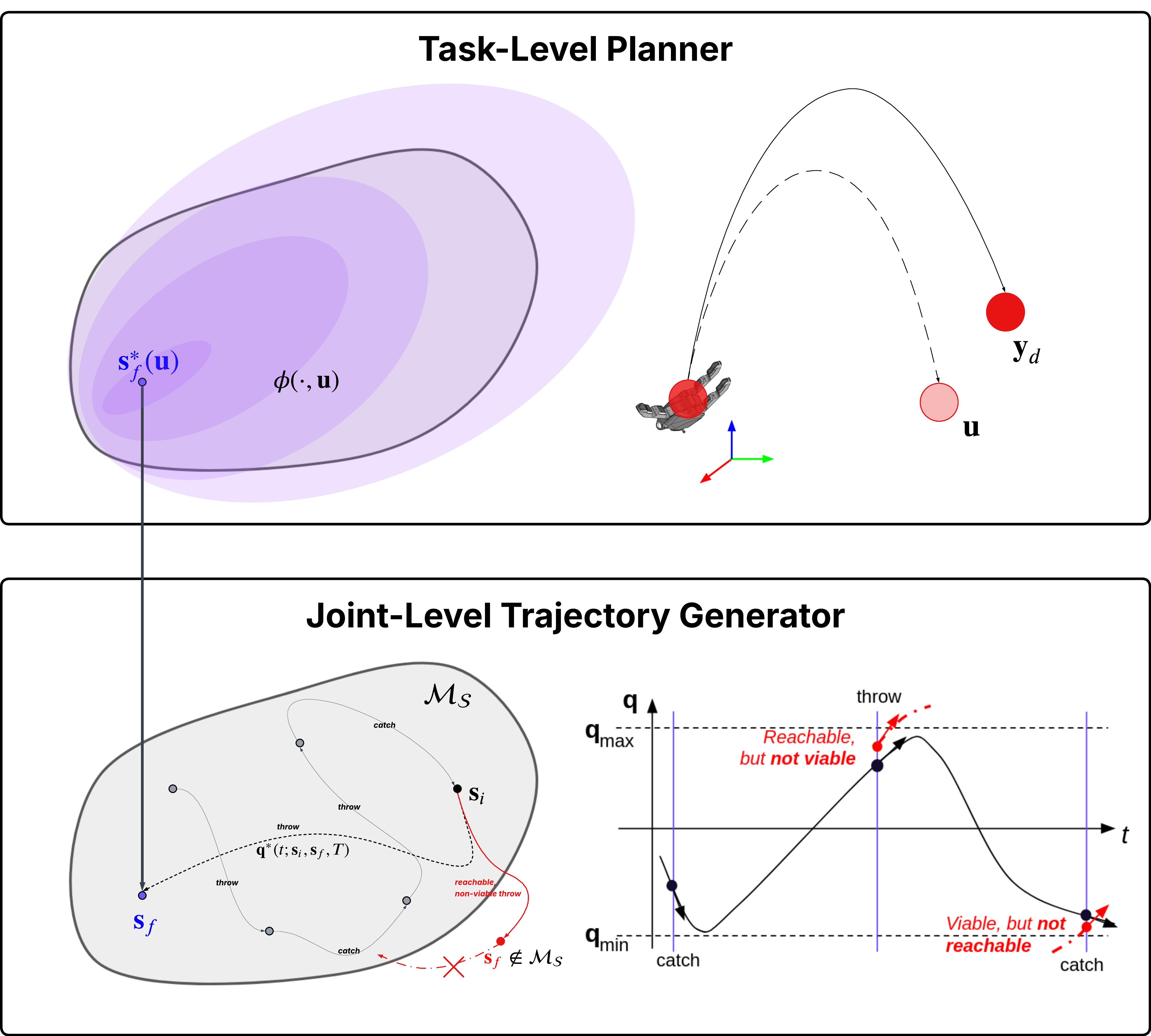}
    \caption{\textbf{Safe dynamic composition of successive skills using the
Mutually Reachable Set ($\mathcal{M}_S$).}
(Top) The Task-Level Planner maps a task command $\mathbf{u}$ from the
Learner to a target joint transition state $\mathbf{s}_f^*$ while
constraining the solution to $\mathcal{M}_S$.
(Bottom left) State-space representation of the Mutually Reachable Set
$\mathcal{M}_S$, illustrating successive transition states within the
set and the feasible trajectories connecting them.
(Bottom right) Illustration of why satisfying instantaneous joint limits
alone is insufficient for safe composition. A state may be reachable
from the current state but leave the joint unable to avoid a subsequent
position limit (\textit{reachable, but not viable}), or may itself allow
safe future motion but not be reachable from the current state without
violating the limits (\textit{viable, but not reachable}).
Restricting transition states to $\mathcal{M}_S$ excludes both cases.}
    \label{fig:control_overview}
\end{figure}

\noindent\textbf{Mutually Reachable Set.}
We define the \textit{Mutually Reachable Set},
$\mathcal{M}_S\subset\mathcal{S}$, such that any two states within
$\mathcal{M}_S$ can be connected in either direction by a feasible
trajectory:
\begin{equation}
\forall\,\mathbf{s}_i,\mathbf{s}_f\in\mathcal{M}_S,\quad
\exists\,\mathbf{q}(\cdot\mid\mathbf{s}_i,\mathbf{s}_f)
\text{ satisfying }\eqref{eq:bc1}\text{--}\eqref{eq:cons2}.
\label{eq:mut_reach_def}
\end{equation}
As a consequence of this bidirectional reachability, any state in $\mathcal{M}_S$ can reach any other state in $\mathcal{M}_S$ through a feasible trajectory, and can likewise be reached from any other state in $\mathcal{M}_S$. Figure~\ref{fig:control_overview}
illustrates this property in terms of reachability and viability \cite{asama2003inev,weiber2008via}.

Constructing the largest $\mathcal{M}_S$ exactly would require reasoning
over the infinite-dimensional space of dynamically feasible trajectories,
which is generally intractable. We therefore restrict the trajectory
space to a finite-dimensional parameterization and construct an inner
approximation of $\mathcal{M}_S$. This parameterization is used only to construct the Mutually Reachable Set offline; trajectories during online execution are still generated using Ruckig as described above. The definition given in (\ref{eq:mut_reach_def}) requires only that a feasible trajectory exists between transition states, rather than that the trajectory generated online use the same parameterization as the offline construction.

For this offline construction, we parameterize trajectories using
B-splines, which allow the position, velocity, acceleration, and jerk
limits to be expressed as finite-dimensional constraints on control parameters. We propagate
forward- and backward-reachable sets numerically from a set of static
states, and their intersection identifies states that can both be reached
from and return to this common set. The resulting set is represented as
a convex polytope,
\begin{equation}
\mathcal{M}_S =
\{\mathbf{s}\mid\mathbf{M}\mathbf{s}\leq\mathbf{m}\}.
\end{equation}
Although this construction may not recover the largest mutually reachable
set, all states retained in $\mathcal{M}_S$ satisfy the mutual-reachability
condition in ~(\ref{eq:mut_reach_def}). Details of the construction and proof are
provided in the Supplementary Materials.

During online execution, the precomputed Mutually Reachable Set is imposed directly on the task-level planner through the linear inequality constraint $\mathbf{M}\mathbf{s}_f \leq \mathbf{m}$. For a task-level command $\mathbf{u}$ from the learner, the planner finds a target transition state $\mathbf{s}_f^*\in\mathcal{M}_S$ by solving
\begin{align}
\label{eq:task_planner}
\mathbf{s}_f^*(\mathbf{u})
= \mathop{\arg\min}_{\mathbf{s}_f}
& \quad \phi(\mathbf{s}_f,\mathbf{u})\\
\mathrm{s.t.}
& \quad \mathbf{g}(\mathbf{s}_f,\mathbf{u})=\mathbf{0},
\nonumber\\
& \quad ~~~~~\mathbf{M}~\mathbf{s}_f \leq \mathbf{m},
\nonumber
\end{align}
where $\phi$ represents task-specific performance objectives and
$\mathbf{g}$ denotes the nonlinear kinematic and task constraints
relating $\mathbf{u}$ to the corresponding throwing or catching
action. Constraining the planner in this way prevents online updates
from producing transition states from which subsequent skills cannot
be feasibly composed. Please refer to the Supplementary Material  (see \eqref{eq:task_planner} and Table~\ref{tab:sup_target_state_optimization}) for the objectives and constraints we used for 
task-level planning of throw and catch skills. 

Importantly, membership in $\mathcal{M}_S$ guarantees that a feasible
trajectory exists between transition states, but does not guarantee
that the transition can be completed within the duration $T$ required
by the orchestrator. This timing requirement is handled separately by
the trajectory generator. If the transition can be completed within
the scheduled duration, Ruckig generates a trajectory with $t_f=T$.
If the robot's kinodynamic limits instead require $t_f>T$, Ruckig
generates the fastest feasible trajectory, but the target state is
reached too late to maintain the prescribed juggling timing. For
example, a ball may be thrown to a catch state that lies within
$\mathcal{M}_S$, yet the receiving arm may not be able to reach that
state before the ball arrives. The resulting delay may lead to a missed
catch, and the system subsequently identifies the juggling failure
when the perception module fails to observe the expected ball from the next throw.

Although we use this particular hierarchical optimization formulation to generate parameterized
low-level behaviors in our experiments, their specific representation
is not fundamental to the framework. As discussed further in the
Discussion, this block could be replaced by other parameterized
behavior representations, such as a goal-conditioned reinforcement
learning policy, provided that the safe composition of successive behaviors can be assured~\cite{kim2023safety}.

\subsection{Regularized Memory-Based Learning for Rapid Adaptation}
\label{sec:memory_learning}
Our learning approach combines physical experience stored in memory
with a prior model of the task-level behavior. For each throw, the
learner first retrieves relevant past experiences, uses them to fit a
local model regularized toward the prior, and then uses this model to
compute a corrected task-level command for the desired outcome. When
relevant experience is unavailable, the local model reduces to the
prior. As experience accumulates, the learned behavior becomes
increasingly determined by physical experience while retaining guidance
from the prior where experience remains limited.

The task-level planner described above provides a nominal mapping from
a task-level command to its expected physical outcome. For throwing,
this mapping assumes an idealized setting in which the rigid ball loses
contact with all fingers instantaneously at the intended release time
and location within the hand, with the intended release velocity, and
subsequently follows a ballistic flight trajectory driven by gravity. We represent this
nominal behavior as a mapping from the current state $\mathbf{x}$ and
command $\mathbf{u}$ to the expected outcome $\mathbf{y}$:
\begin{equation}
    \mathbf{y} = \mathbf{f}_0(\mathbf{x},\mathbf{u}),
\end{equation}
where $\mathbf{f}_0$ represents the prior model available before
on-robot learning. For the throw skill used in our juggling experiments, we define
\begin{equation}
\mathbf{x}
\triangleq
(\mathbf{p}_{\mathrm{catch}}^{-},\mathbf{p}_{\mathrm{land}}^{-})
\in\mathbb{R}^{6},
~~
\mathbf{u}
\triangleq
\mathbf{p}_{\mathrm{land}}^{u}
\in\mathbb{R}^{3},
~~
\mathbf{y}
\triangleq
\mathbf{p}_{\mathrm{land}}
\in\mathbb{R}^{3}.
\end{equation}
Here, $\mathbf{p}_{\mathrm{catch}}^{-}$ is the commanded catch position
for the preceding ball,
and $\mathbf{p}_{\mathrm{land}}^{-}$ is the latest estimate of the ball's
landing position available at the start of the subsequent
throw. The two quantities differ because additional visual
observations become available after the catch command has been fixed.
We include both because their difference provides useful information
about how the ball is received by the hand. This difference can be
particularly large for the open-loop catches of side-thrown balls in
the box and shower patterns. The command
$\mathbf{p}_{\mathrm{land}}^{u}$ specifies the desired landing
position of the current throw, and $\mathbf{p}_{\mathrm{land}}$
is its observed physical outcome.

Because the nominal throwing behavior is generated to throw the
ball to $\mathbf{p}_{\mathrm{land}}^{u}$, the corresponding prior
task-level model is simply an identity mapping:
\begin{equation}
    \mathbf{f}_0(\mathbf{x},\mathbf{u})=\mathbf{u}.
\end{equation}
More generally, for a goal-conditioned behavior in which the command directly specifies the desired task outcome, the prior can take the same identity form. For other parameterized behaviors, $\mathbf{f}_0$ can instead be obtained from an analytical model, simulation, or offline learning. Without adaptation, the command for a desired outcome $\mathbf{y}_d$ would be determined solely from the prior model as
\begin{align}
\mathbf{u}^* = \arg\min_{\mathbf{u}}& \|\mathbf{y} - \mathbf{y}_d\|^2 \\
\text{s.t.}&~ \mathbf{y} = \mathbf{f}_0(\mathbf{x},\mathbf{u}).
\end{align}
For the identity prior used here, this simply gives
$\mathbf{u}^*=\mathbf{y}_d$. On the physical robot, however, the
resulting outcome $\mathbf{f}(\mathbf{x},\mathbf{u}^*)$ can differ
substantially from $\mathbf{y}_d$. In our experiments, this discrepancy
was large enough that using the prior alone could not complete even
a single cycle of juggling.

This discrepancy arises from complex physical interactions that are
not represented by the nominal throwing behavior. Variations in
multi-finger contact timing, friction, grasp alignment, and ball
slippage can cause the resulting throw to deviate substantially from
its nominal outcome. The actual task-level relationship between the
state, command, and physical outcome can therefore be written as
\begin{equation}
\mathbf{y} = \mathbf{f}(\mathbf{x}, \mathbf{u}),
\label{eq:task_dynamics}
\end{equation}
which can deviate significantly from the prior $\mathbf{f}_0$. 

To correct this discrepancy online, we use a regularized memory-based learning approach that combines on-robot experience with the prior model $\mathbf{f}_0$. Importantly, $\mathbf{f}_0$ does not need to accurately predict the task outcome to remain useful for learning. Its first-order gradient can still provide useful information about how changes in the state and command affect the outcome locally, even when its absolute prediction is inaccurate. 


Given a target outcome $\mathbf{y}_d$ and the current state $\mathbf{x}$, we retrieve the $k$-nearest neighbors ($k\text{-NN}$) from a memory buffer
\begin{equation*}
\mathcal{D} = \{(\mathbf{x}_i, \mathbf{u}_i, \mathbf{y}_i)\}_{i=1}^N.
\end{equation*}
Because the command required to achieve $\mathbf{y}_d$ depends on both the current state and the desired outcome, neighbors are retrieved in the joint $(\mathbf{x},\mathbf{y})$ space. We assign a weight $w_i$ to each neighbor using a localized Radial Basis Function (RBF) kernel:
\begin{equation*}
w_i = k\big((\mathbf{x}_i, \mathbf{y}_i), (\mathbf{x}, \mathbf{y}_d)\big) = \exp\left( -\left( \frac{\|\mathbf{x}_i - \mathbf{x}\|^2}{h_x^2} + \frac{\|\mathbf{y}_i - \mathbf{y}_d\|^2}{h_y^2} \right) \right)
\end{equation*}
where the bandwidth parameters $h_x$ and $h_y$ are fixed scaling constants selected to reflect the empirical scale of variations in $\mathbf{x}$ and $\mathbf{y}$, respectively. The nominal reference task command $\bar{\mathbf{u}}$ is then computed as the localized weighted average of the retrieved memory:
\begin{equation}
\bar{\mathbf{u}}(\mathbf{x},\mathbf{y}_d) = \frac{\sum_{i=1}^{k_{\text{NN}}} w_i \cdot\mathbf{u}_i}{\sum_{i=1}^{k_{\text{NN}}} w_i}.
\end{equation}
While this Nadaraya--Watson estimator is effective for interpolation when the query lies within a sufficiently dense region of the data, its estimate becomes unreliable when relevant experience is sparse~\cite{atkeson1997locally}. This is particularly important during the initial phase of adaptation, when the robot must improve beyond the limited behaviors it has already experienced. The prior model $\mathbf{f}_0$ provides first-order information that can guide this necessary extrapolation.

To connect the non-parametric memory lookup to the prior model, we incorporate $\mathbf{f}_0$ into a local linear regression. We identify the local relationship between state, command, and outcome by regularizing a weighted least-squares regression toward the prior. The local Jacobians $\mathbf{C}, \mathbf{D}$ and offset $\mathbf{d}$ around the current operating point are found by solving:
\begin{align}
\min_{\mathbf{C}, \mathbf{D}, \mathbf{d}} \sum_{i=1}^{k_{\text{NN}}} w_i &\|\mathbf{y}_i - \mathbf{C} \delta\mathbf{x}_i - \mathbf{D} \delta\mathbf{u}_i - \mathbf{d}\|^2 + \gamma \Bigg[ \left\|\mathbf{C} - \frac{\partial \mathbf{f}_0}{\partial \mathbf{x}} (\mathbf{x}, \bar{\mathbf{u}}) \right\|_F^2  \nonumber\\
&+ \left\|\mathbf{D} - \frac{\partial \mathbf{f}_0}{\partial \mathbf{u}}(\mathbf{x}, \bar{\mathbf{u}}) \right\|_F^2 + \| \mathbf{d} - \mathbf{f}_0(\mathbf{x}, \bar{\mathbf{u}})\|^2 \Bigg] \label{eq:reg_local_regression}
\end{align}
where $\delta\mathbf{x}_i = \mathbf{x}_i - \mathbf{x}$ and $\delta\mathbf{u}_i = \mathbf{u}_i - \bar{\mathbf{u}}$, $\|\cdot\|_F$ denotes the Frobenius norm, and $\gamma > 0$ is a regularization hyperparameter balancing the local physical experience against the prior (see Figure~\ref{fig:learning_curve_gamma}). For the identity prior used in our experiments, where $\mathbf{f}_0(\mathbf{x}, \mathbf{u}) \equiv \mathbf{u}$, the prior derivatives simplify to $\frac{\partial \mathbf{f}_0}{\partial \mathbf{u}} \equiv \mathbf{I}$ and $\frac{\partial \mathbf{f}_0}{\partial \mathbf{x}} \equiv \mathbf{0}$. Thus, where physical experience is limited, the learner retains the simple prior that a change in the commanded landing position produces the same change in the actual landing position. As experience accumulates, the locally fitted relationship can depart from this prior according to the behavior observed on the physical robot. The above quadratic optimization problem has a closed-form solution derived in the Supplementary Materials (see Equation~\ref{eq:theta_star_cf}).

Once this local linear model $(\mathbf{C}^*, \mathbf{D}^*, \mathbf{d}^*)$ is identified, the task command could be obtained by directly inverting the local model, i.e., $\mathbf{u}^* = \bar{\mathbf{u}} + {\mathbf{D}^*}^{-1}(\mathbf{y}_d - \mathbf{d}^*)$. In practice, however, direct inversion can produce excessively large command changes when the local approximation is imperfect. To obtain more stable adaptation, the optimal command $\mathbf{u}^*$ is instead computed by solving a regularized quadratic optimization problem:
\begin{align}
\mathbf{u}^* &= \arg\min_{\mathbf{u}} \|\mathbf{y} - \mathbf{y}_d \|^2 + \eta \|\mathbf{u} - \bar{\mathbf{u}}\|^2 \\
             & \quad\quad\quad \text{s.t. } \quad \mathbf{y} = \mathbf{D}^* (\mathbf{u} - \bar{\mathbf{u}}) + \mathbf{d}^* \\ 
             &= \bar{\mathbf{u}} + \left({\mathbf{D}^*}^\top\mathbf{D}^* + \eta \mathbf{I}\right)^{-1}\left[{\mathbf{D}^*}^\top (\mathbf{y}_d - \mathbf{d}^*)\right] \label{eq:u_closed_form}
\end{align}
where $\eta > 0$ penalizes large adaptation steps and keeps the command near the locally weighted reference $\bar{\mathbf{u}}$. A fixed value of $\eta = 0.3$ is used for all experiments in this work. Equation~\ref{eq:u_closed_form} has the same form as a Levenberg--Marquardt update, with the estimated local gradient $\mathbf{D}^*$ playing the role of the Jacobian and the regularization balancing the Gauss--Newton and gradient-descent regimes~\cite{marquardt1963algorithm}.

We note that the resulting $\mathbf{u}^*(\mathbf{x}, \mathbf{y}_d, \mathcal{D})$ defines a closed-loop, memory-based policy, although the state $\mathbf{x}$ and the corresponding local Jacobian $\mathbf{C}^*$ do not explicitly appear in Equation~\ref{eq:u_closed_form}. Their influence enters through the memory retrieval and local model fitting: the current state $\mathbf{x}$ determines which experiences are considered relevant and therefore affects both $\bar{\mathbf{u}}$ and the locally fitted model used to compute the command.

Lastly, the choice of task-level parameterization, distance metric, and kernel bandwidth determines how experiences are weighted in the memory-based learning problem. In our formulation, $\mathbf{u}$ and $\mathbf{y}$ have the same physical interpretation as the commanded and observed ball landing positions, respectively, yielding the identity prior $\mathbf{f}_0(\mathbf{x},\mathbf{u})=\mathbf{u}$. Because memory retrieval and regularization depend on distances and norms in the chosen coordinates, an alternative parameterization can change which experiences are considered similar. For example, parameterizing the same throwing behavior by release velocity would produce different distances between experiences despite representing the same physical action. Under a linear reparameterization, an equivalent formulation can be recovered by appropriately transforming the distance metrics used by the kernel and replacing the Euclidean and Frobenius norms in the regularization with corresponding weighted norms. Rather than selecting the metric and bandwidth solely from the numerical scales of the chosen coordinates, they can also be selected or learned from the distribution of experience and the local variation of the input-output relationship~\cite{atkeson1997locally,noh2010generative,noh2017metric}.

\subsection{Perception}
\label{sec:perception}

Our perception system is designed to provide sufficiently accurate object pose estimates at low latency while remaining general enough to accommodate new objects without task-specific data collection or retraining. To achieve this, the system processes synchronized RGB-D data through a modular pipeline that combines prompt-driven image segmentation with CAD-based geometric tracking. While a dedicated lightweight model trained specifically for a particular object and operating condition could provide excellent speed and accuracy for high-speed juggling, changing the object would then require collecting new data and retraining the model. Instead, our pipeline introduces a new object using only a few point prompts on a handful of frames together with its CAD geometry.

For the instance segmentation phase, we utilize the Efficient Track Anything Model (EfficientTAM) to maintain maximum flexibility when introducing new objects \cite{xiong2025efficient}. A critical advantage of this choice is that EfficientTAM is relatively lightweight compared to standard foundational models; when deployed on an NVIDIA RTX A6000 GPU, it achieves an inference time of just 0.01 second. Instead of a dedicated training process, the module is initialized using four calibration images, in which the user selects a few image points on each target object (e.g., each juggling ball) to indicate which objects should be segmented. These user-selected points serve as the point prompts to EfficientTAM. During execution, the network processes raw images to output dense instance masks bounding each individual ball. If a user decides to switch to a different set of balls, the same initialization can be repeated by selecting a few points on the new objects, bypassing any task-specific data collection or training.

The calculated masks are subsequently used to segment the 3D point cloud from the depth camera. Using the calibrated extrinsic transformation between the depth and RGB cameras and the RGB camera intrinsics, each 3D point is projected onto the RGB image plane, and points whose projections fall within each object mask are retained to obtain the segmented 3D point cloud of the corresponding object. To estimate the current 3D position of each object at minimal computational cost, we perform scan-to-model registration using Fast Generalized Iterative Closest Point (FastGICP) \cite{small_gicp}. Given a standard CAD model of the object geometry (a sphere in this configuration), FastGICP aligns the reference model cloud to the observed point cloud to extract the rigid body translation, yielding the tracked 3D coordinate of the ball.

As consecutive 3D positions are gathered during a single flight phase, they are aggregated within a rolling buffer to estimate the object's airborne state. Assuming unperturbed ballistic flight under gravity, the system fits a parabolic trajectory to the accumulated observations and predicts the ball position at the absolute future time specified by the orchestrator. This forward-predicted point serves as the target catching state passed directly to the task planner.

The same perception pipeline is intended to remain applicable when the manipulated object changes. Because object identity is introduced through point prompts and object geometry through a CAD model, adapting the pipeline does not require retraining the perception network. For a non-spherical object such as a juggling club, the same pipeline can be extended to estimate both position and orientation using the corresponding CAD geometry, while the simple ballistic prediction used for spherical balls can be extended to account for the club's rotational dynamics using its approximately axisymmetric inertia.

\section*{Acknowledgments}
We thank the members of the APEX team, Gregory Xie, Annan Mozeika, Joseph Aronov, and Christopher Keeley, for their  maintenance of the hardware platform throughout our experiments. We also thank all members of the CONDOR team for their software infrastructure maintenance and constructive discussions throughout this project.

\textbf{Funding:}
This research was supported by the CONDOR and APEX teams at the Robotics and AI (RAI) Institute, with funding provided by the Hyundai Motor Group.

\textbf{Author contributions:}
T.L. led the research and development of the overall framework, including the learning, planning, and perception modules of the system. T.L. and C.W. co-implemented the framework, and designed and conducted the physical hardware experiments. C.G.A. provided technical consultation and insights on memory-based learning throughout the project. A.A.R. and N.R. initiated and directed the project on robot juggling.

\textbf{Competing interests:}
The authors declare that they have no competing interests.

\textbf{Data and materials availability:}
All data needed to evaluate the conclusions in the paper are present in the paper or the Supplementary Materials.

\bibliography{section/reference} 

\clearpage\newpage


\renewcommand{\thefigure}{S\arabic{figure}}
\renewcommand{\themovie}{S\arabic{movie}}
\renewcommand{\thetable}{S\arabic{table}}
\renewcommand{\theequation}{S\arabic{equation}}
\renewcommand{\thepage}{S\arabic{page}}
\setcounter{figure}{0}
\setcounter{table}{0}
\setcounter{equation}{0}
\setcounter{movie}{0}
\setcounter{page}{1} 

\def\scititle{Rapid On-Robot Learning for Dynamic Manipulation Skills: Robot Juggling}

\section*{Supplementary Materials for\\ \scititle}

\author{Taeyoon Lee}, 
\author{Chunpeng Wang}, 
\author{Christopher G. Atkeson}, 
\author{Alfred A. Rizzi}, and 
\author{Nicolas Rojas}

\subsubsection*{This PDF file includes:}

\begin{itemize}
    \setlength{\itemsep}{0pt}
    \setlength{\parskip}{0pt}
    \item Fig S1.
Shared target positions for juggling skills
    \item Fig S2.
    Mutually reachable set construction for the right arm joints ($\text{RJ1}$--$\text{RJ7}$) of AthenaZero.
    \item Table S1.
Coordinates of the shared target positions for juggling skills
    \item Table S2. Optimization formulations of the task-level planner for throwing and catching tasks
    \item Table S3. Skill sequence configuration for the cascade juggling pattern
    \item Table S4. Skill sequence configuration for continuous tennis, half shower, and cascade sequence of patterns
    \item Table S5. Skill sequence configuration for the shower juggling pattern
    \item Table S6. Skill sequence configuration for the box juggling pattern
\end{itemize}

\subsubsection*{Other Supplementary Materials for this manuscript:}
\begin{itemize}
    \setlength{\itemsep}{0pt}
    \setlength{\parskip}{0pt}
    \item Movie S1.
Learning cascade juggling pattern
    \item Movie S2.
Learning continuous tennis, half shower, and cascade patterns
    \item Movie S3.
Learning shower juggling pattern
    \item Movie S4.
Learning box juggling pattern
\end{itemize}


\subsection*{Materials and Methods}

\subsubsection*{Skill description}

As summarized in Table~\ref{tab:skill_param}, each juggling skill is fully parameterized by a tuple containing its {type}, {object ID}, {task identifier}, and {duration}. 

The {type} and {task identifier} function as semantic routing keys that specify the task-level objectives, directly altering the optimization formulations within the downstream learning and planning nodes. Every \texttt{throw} skill is mapped to a task identifier string formatted as \texttt{to\{XX\}from\{YY\}}, where the indices \texttt{XX} and \texttt{YY} resolve to the target ball landing position $\textbf{p}_{\mathrm{land}}$ and the target release position $\mathbf{p}_{\mathrm{release}}$, respectively. Conversely, a \texttt{catch} skill is formatted as \texttt{at\{XX\}from\{YY\}}, where \texttt{XX} denotes a nominal catching location $\textbf{p}_{\mathrm{catch}}$ (which is continuously overridden online by real-time perception updates) and \texttt{YY} denotes the origin of the ball's ballistic flight. Beyond these positions, each skill encapsulates supplementary execution parameters, including finger open/close position commands and the exact timing offsets used to trigger them near the end of execution. The explicit incorporation of these parameters within the learner and task-level planner optimizations is detailed in the subsequent sections of this document.

\begin{figure}[t]
	\centering
	\includegraphics[width=\linewidth]{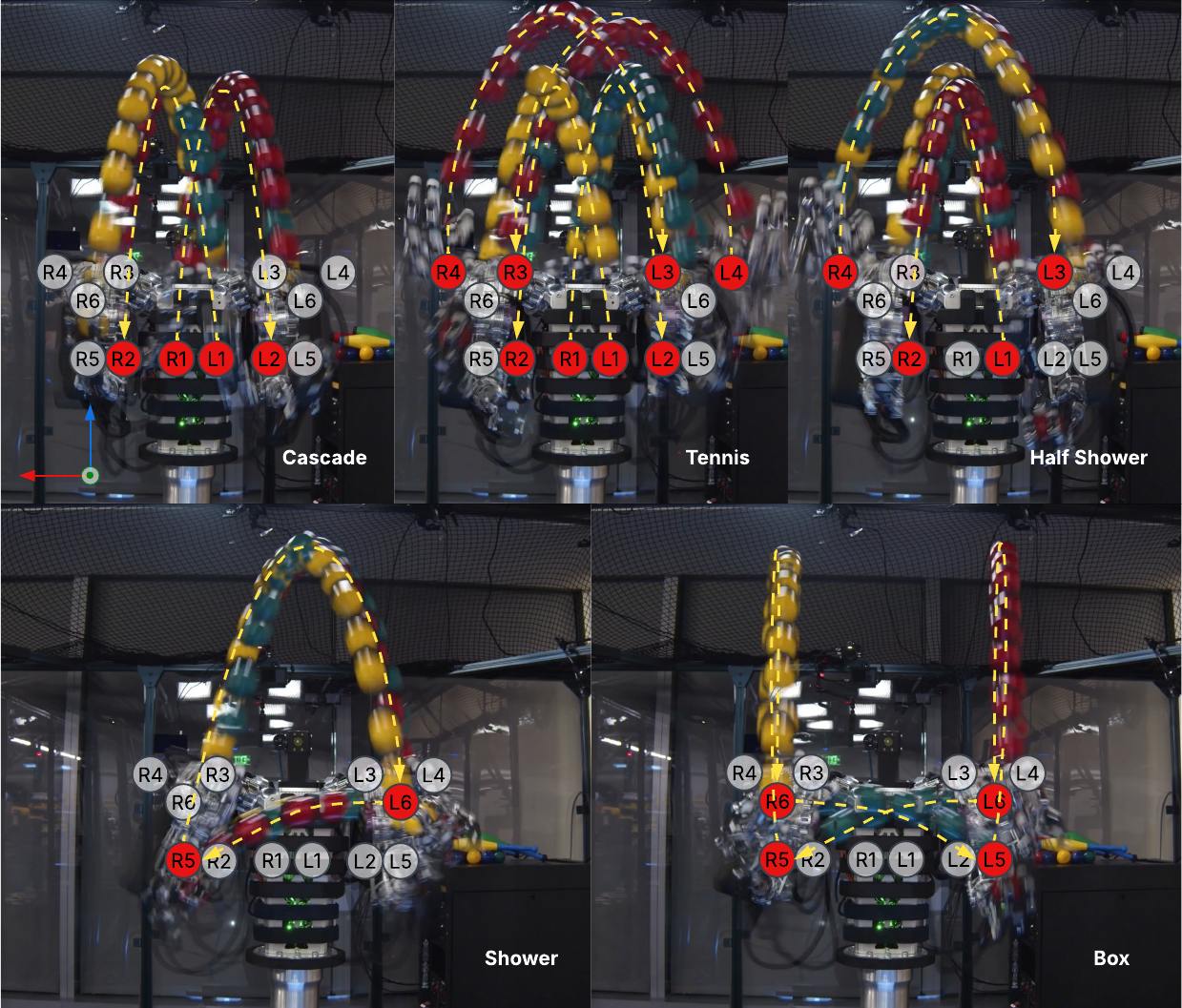}
	\caption{\textbf{Shared target positions for juggling skills.}
            The six designated spatial targets for the right arm ($R1$–$R6$) and left arm ($L1$–$L6$) are shared across the target release, target land/catch configurations. Active targets utilized in a specific juggling pattern are highlighted in red (e.g., $R1$, $R2$, $L1$, and $L2$ are the enabled positions for the cascade pattern).}
	\label{fig:sup_skill_pos}
\end{figure}

The manual selection of the target operational positions ($\mathbf{p}_{\mathrm{release}}$, $\textbf{p}_{\mathrm{land}}$, and $\textbf{p}_{\mathrm{catch}}$) for each skill is subject to several kinodynamic and perceptual constraints. Spatially, these waypoints must remain sufficiently clustered to ensure the joint-level trajectory generator can resolve feasible paths within the execution window, yet remain sufficiently separated to maintain a safe physical margin that prevents mid-air ball collisions. Perceptually, the ballistic parabolas resulting from these coordinates must reside within the egocentric field-of-view of the torso-mounted RGB-D camera suite to ensure continuous visual tracking. In accordance with these constraints, we establish a set of 12 shared operational positions, as illustrated in Figure~\ref{fig:sup_skill_pos} and tabulated in Table~\ref{tab:sup_skill_pos}; these continuous coordinate values are indexed and retrieved dynamically during real-time parsing of the task identifier string.

The {object ID} and {duration} parameters primarily govern asynchronous node coordination between the skill execution and perception nodes. Upon receiving a completion callback from a \texttt{throw} skill, the orchestrator dispatches a tracking request for the corresponding {object ID}, passing along the nominal flight duration. Upon receiving this asynchronous request, the perception node initiates tracking for that specific ball and continuously streams its predicted landing position at the future timestamp dictated by the provided flight duration. Concurrently, the {duration} parameter is used by the skill executor block, where it is utilized both to define the task-level throw velocity command (as detailed in Table~\ref{tab:sup_target_state_optimization}) and to scale the underlying joint-level trajectory generator to match the allocated execution window. As outlined in Section~2\ref{sec:sys_overview}, rather than passing fixed nominal duration intervals directly to these distributed nodes, the system converts all object tracking and skill execution duration into absolute targets mapped to a global wall-clock time, thereby preventing cumulative phase drift of the overall bimanual juggling pattern.

\subsubsection*{Juggling pattern}

Every juggling pattern has three consecutive phases: \textit{Start}, \textit{Repeat}, and \textit{Stop}. The \textit{Start} phase initializes the pattern with one hand holding two objects and the opposing hand holding one object. The \textit{Repeat} phase comprises the periodic sequence of coordinated throws and catches that defines the specific juggling pattern. Finally, the \textit{Stop} phase safely terminates execution by capturing two objects within a single hand and securing the third object with the other hand. Individual skill descriptions are structured uniformly in the format \texttt{[Type, Object ID, Task Identifier, Duration]}. The complete sequence of skill descriptions for the single cascade, the tennis $\rightarrow$ half-shower $\rightarrow$ cascade, the shower, and the box patterns are presented in Table~\ref{tab:sup_pattern_cascade}, \ref{tab:sup_pattern_tennis_halfshower_cascade}, \ref{tab:sup_pattern_shower}, and \ref{tab:sup_pattern_box}.


\begin{table}[t]
	\centering
    \rowcolors{2}{white}{gray!15}
	\caption{\textbf{Coordinates of the shared target positions for juggling skills.} The $x$-, $y$-, and $z$-axes correspond to the rightward, forward, and upward directions relative to the robot base, respectively.}
	\label{tab:sup_skill_pos}
	\begin{tabular}{c c c c | c c c c}
		\hline
		\textbf{ID} & \textbf{$x$ (m)} & \textbf{$y$ (m)} & \textbf{$z$ (m)} & \textbf{ID} & \textbf{$x$ (m)} & \textbf{$y$ (m)} & \textbf{$z$ (m)} \\
		\hline
		R1 & \phantom{-}0.056 & 0.415 & 1.35  & L1 & -0.056           & 0.415 & 1.35\\
		R2 & \phantom{-}0.206 & 0.415 & 1.35  & L2 & -0.206           & 0.415 & 1.35\\
		R3 & \phantom{-}0.206 & 0.415 & 1.50  & L3 & -0.206           & 0.415 & 1.50\\
		R4 & \phantom{-}0.356 & 0.415 & 1.50  & L4 & -0.356           & 0.415 & 1.50\\
		R5 & \phantom{-}0.226 & 0.415 & 1.35  & L5 & -0.226           & 0.415 & 1.35\\
		R6 & \phantom{-}0.226 & 0.415 & 1.45  & L6 & -0.226           & 0.415 & 1.45\\
		\hline
	\end{tabular}
\end{table}

\subsubsection*{Optimization formulation of the task-level planner}

Given a task-level command $\mathbf{u}$, the Task-Level Planner
determines the target joint transition state $\mathbf{s}_f$ by solving
the task-specific constrained optimization summarized in
Table~\ref{tab:sup_target_state_optimization}. The constraints encode
the task-space conditions required for throwing or catching, while the
objective selects a preferred joint state when multiple solutions
satisfy these conditions. In both cases, the target state is additionally
constrained to the Mutually Reachable Set so that the resulting skill
can be safely composed with subsequent skills.

For a \texttt{catch} skill, the task-level command
$\mathbf{p}_{\mathrm{catch}}$ specifies the predicted position at which
the incoming ball should be intercepted. The target velocity and
acceleration are set to zero, such that the hand comes to rest at the
end of the catch. The forward-kinematics constraint places the hand at
the commanded catch position, while the orientation constraint requires
the palm normal to remain within $\theta_{\max}$ of the desired catch
orientation. The objective selects among the remaining inverse-kinematic
solutions by penalizing joint displacement while favoring alignment of
the palm with the incoming ball.

For a \texttt{throw} skill, the task-level command
$\mathbf{p}_{\mathrm{land}}^{u}$ specifies the commanded landing
position of the ball. The planner optimizes the complete transition
state, including joint position, velocity, and acceleration. The
release position is determined by the forward kinematics, while the
required release velocity is computed from the commanded landing
position and prescribed flight duration under ballistic flight. This
Cartesian release velocity is then related to the joint velocity
through the robot Jacobian. The palm orientation is also constrained
to align approximately with the release-velocity direction.

The acceleration constraint is introduced to facilitate clean ball release. At the intended release instant, the ball transitions from moving with the hand to free flight, during which its acceleration is g. We therefore constrain the hand acceleration at this instant to also equal g. Together with the prescribed release velocity, this makes the relative velocity and acceleration between the hand and the nominal ballistic motion zero at the release boundary, reducing additional contact forces as the fingers open. The objective then resolves the remaining redundancy by favoring smaller joint position, velocity, and acceleration while aligning the palm with the throw direction.

Together, these constraints represent the idealized throwing assumption
used by the Task-Level Planner: the ball separates from all fingers at
the planned release time and location with the planned velocity and
then follows ballistic flight under gravity. In the physical system,
however, finger--ball contact does not terminate instantaneously, and
friction, slippage, actuation, and trajectory-tracking errors cause the
realized throw to deviate from this idealized behavior. The Regularized
Memory-Based Learner compensates for this mismatch by adapting the
task-level landing command as described in
Section~4\ref{sec:memory_learning}.

The complete objectives, constraints, and fixed quantities for both
skills are summarized in
Table~\ref{tab:sup_target_state_optimization}. We solve the resulting
nonlinear constrained optimization problems using the PSQP sequential
quadratic programming solver implemented in
\texttt{pyOptSparse}~\cite{wu2020pyoptsparse}. The optimization
typically converges to a feasible solution within approximately
$10$--$20\,\mathrm{ms}$.

\subsubsection*{Robot joint kinematic and dynamic limit specifications}
The kinematic joint position limits, defined by the lower bound $\mathbf{q}_{\mathrm{min}}$ and upper bound $\mathbf{q}_{\mathrm{max}}$, are derived directly from the physical hard stops of each custom actuator unit. The joint velocity limits $\mathbf{v}_{\mathrm{max}}$, due to maximum drive voltages and back-EMF, are determined analytically from the motor specifications and the custom planetary gear box reduction ratios. 

The joint acceleration limits $\mathbf{a}_{\mathrm{max}}$ are fundamentally bounded by the maximum peak torque capabilities $\boldsymbol{\tau}_{\mathrm{max}}$ of the actuator units. The transformation from joint-space torque to acceleration is governed by the standard rigid-body manipulator dynamics:
\begin{equation}
    \mathbf{M}(\mathbf{q})\ddot{\mathbf{q}} + \mathbf{C}(\mathbf{q}, \dot{\mathbf{q}})\dot{\mathbf{q}} + \mathbf{g}(\mathbf{q}) = \boldsymbol{\tau}
\end{equation}
where $\mathbf{M}(\mathbf{q})$ denotes the joint-space inertia matrix, $\mathbf{C}(\mathbf{q}, \dot{\mathbf{q}})$ represents the Coriolis and centrifugal matrix, and $\mathbf{g}(\mathbf{q})$ is the gravity vector. Although the true instantaneous acceleration limit is coupled and state-dependent, we establish a conservative, constant acceleration bound by evaluating the system dynamics at a nominal, task-relevant workspace configuration $\mathbf{q}_c \in \mathbb{R}^n$ where the primary throwing and catching phases occur. Furthermore, neglecting the velocity-dependent Coriolis terms and assuming decoupled joint inertia yields the decoupled upper and lower acceleration bounds:
\begin{equation}
\begin{aligned}
\mathbf{a}_{\mathrm{upper}} &= r_a \operatorname{diag}\left(\mathbf{M}(\mathbf{q}_c)^{-1}\right)\left(\boldsymbol{\tau}_{\mathrm{max}} - \mathbf{g}(\mathbf{q}_c)\right) \\
\mathbf{a}_{\mathrm{lower}} &= r_a \operatorname{diag}\left(\mathbf{M}(\mathbf{q}_c)^{-1}\right)\left(-\boldsymbol{\tau}_{\mathrm{max}} - \mathbf{g}(\mathbf{q}_c)\right)
\end{aligned}
\end{equation}
where $\operatorname{diag}(\cdot)$ extracts the diagonal elements of the inverted inertia matrix. To enforce a symmetric bound, the final operational limit $\mathbf{a}_{\mathrm{max}}$ is defined element-wise as:
\begin{equation}
    \mathbf{a}_{\mathrm{max}} = \min\left(\mathbf{a}_{\mathrm{upper}},\, -\mathbf{a}_{\mathrm{lower}}\right)
\end{equation}
Here, $r_a \in (0, 1]$ is an empirical safety factor introduced to provide a conservative margin against unmodeled dynamics and transient peaks during high-speed trajectories. In this work, selecting $r_a = 0.8$ was determined to successfully prevent actuator saturation during juggling behaviors.

The joint jerk limit $\mathbf{j}_{\mathrm{max}}$, representing the maximum allowable rate of change of acceleration, is similarly mapped from the actuator torque rate limit $\dot{\boldsymbol{\tau}}_{\mathrm{max}}$. For our custom actuator units, this underlying torque derivative limit is governed by the electrical time constant of the motor windings and the bandwidth of the current loop controller. We identify this limit experimentally via closed-loop frequency response identification, defining the maximum frequency at which the motor current can track a sinusoidal reference. By taking the time derivative of the simplified manipulator dynamics at $\mathbf{q}_c$, dropping the state-dependent derivatives ($\dot{\mathbf{M}}$, $\dot{\mathbf{g}}$), and treating the joint axes as decoupled, the operational jerk limit is approximated as:
\begin{equation}
    \mathbf{j}_{\mathrm{max}} = r_j \operatorname{diag}\left(\mathbf{M}(\mathbf{q}_c)^{-1}\right)\dot{\boldsymbol{\tau}}_{\mathrm{max}}
\end{equation}
where $r_j \in (0, 1]$ is a corresponding empirical safety coefficient introduced to account for the omitted higher-order derivatives. We set $r_j = 0.8$ to ensure reliable, un-saturated trajectory tracking in practice.

\begin{figure*}[h!]
    \centering
    \includegraphics[width=\linewidth]{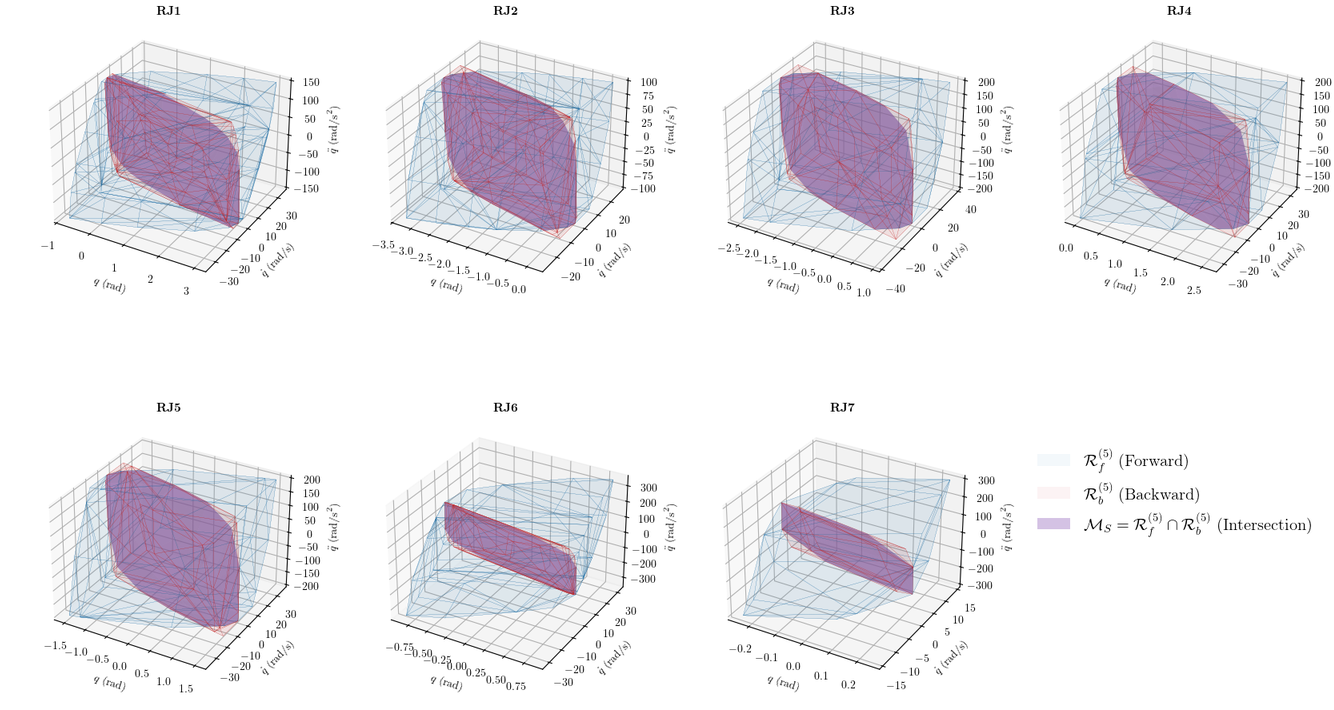}
    \caption{\textbf{Mutually reachable set construction for the right arm joints ($\text{RJ1}$--$\text{RJ7}$) of AthenaZero.} 
Each subplot displays the overlay of the forward reachable set $\mathcal{R}_f^{(5)}$ (light blue), backward reachable set $\mathcal{R}_b^{(5)}$ (light red), and their resulting intersection—the mutually reachable set $\mathcal{M}_S = \mathcal{R}_f^{(5)} \cap \mathcal{R}_b^{(5)}$ (purple). The volumes are mapped within the joint phase-space coordinates of position $q$, velocity $\dot{q}$, and acceleration $\ddot{q}$, illustrating the dynamically feasible target joint state constraint for each independent degree of freedom.}
    \label{fig:mut_reach}
\end{figure*}

\subsubsection*{Numerical polytope approximation of the mutually reachable set}

Over a local time horizon $t\in[0,\tau]$, the trajectory of a single
joint is represented as
\begin{equation}
q(t;\mathbf{p})
=
\sum_{j=1}^{n_c} B_{j,d}(t)p_j
=
\mathbf{B}_{d}(t)\mathbf{p},
\end{equation}
where $\mathbf{p}\in\mathbb{R}^{n_c}$ contains the B-spline control
points and $\mathbf{B}_{d}(t)\in\mathbb{R}^{1\times n_c}$ contains the
degree-$d$ B-spline basis functions. Because B-splines are closed under
differentiation, the $r$th derivative can be written as
\begin{equation}
q^{(r)}(t)
=
\mathbf{B}_{d-r}(t)\mathbf{D}^{(r)}\mathbf{p},
\end{equation}
where
\begin{equation}
\mathbf{D}^{(r)}
\in
\mathbb{R}^{(n_c-r)\times n_c},
\qquad
\mathbf{B}_{d-r}(t)
\in
\mathbb{R}^{1\times(n_c-r)},
\end{equation}
and $\mathbf{D}^{(r)}$ maps the original control points to the control
points of the $r$th derivative. We define
$\mathbf{D}^{(0)}=\mathbf{I}_{n_c}$ as a $n_c \times n_c$ identity matrix.

By the convex-hull property of B-splines, bounding the control points
of each derivative provides sufficient conditions for satisfying the
corresponding continuous-time limits over the entire interval. We
therefore impose
\begin{align}
q_{\min}\cdot\mathbf{1}_{n_c}
&\leq
\mathbf{D}^{(0)}\mathbf{p}
\leq
q_{\max}\cdot\mathbf{1}_{n_c}, \nonumber \\
-v_{\max}\cdot\mathbf{1}_{n_c-1}
&\leq
\mathbf{D}^{(1)}\mathbf{p}
\leq
v_{\max}\cdot\mathbf{1}_{n_c-1},
\nonumber\\
-a_{\max}\cdot\mathbf{1}_{n_c-2}
&\leq
\mathbf{D}^{(2)}\mathbf{p}
\leq
a_{\max}\cdot\mathbf{1}_{n_c-2}, \nonumber \\
-j_{\max}\cdot\mathbf{1}_{n_c-3}
&\leq
\mathbf{D}^{(3)}\mathbf{p}
\leq
j_{\max}\cdot\mathbf{1}_{n_c-3},
\end{align}
where $\mathbf{1}_m\in\mathbb{R}^{m}$ denotes the vector of ones.
Stacking these inequalities gives the polyhedral set of feasible
B-spline parameters
\begin{equation}
\mathcal{P}_{\mathrm{feasible}}
=
\left\{
\mathbf{p}\in\mathbb{R}^{n_c}
\mid
\mathbf{D}_c\mathbf{p}\leq\mathbf{b}_c
\right\}.
\end{equation}

The transition states at the beginning and end of each interval are
linear functions of the same control points:
\begin{equation}
\mathbf{s}(0)
=
\mathbf{D}_0\mathbf{p},
\qquad
\mathbf{s}(\tau)
=
\mathbf{D}_f\mathbf{p},
\end{equation}
where $\mathbf{D}_0,\mathbf{D}_f\in\mathbb{R}^{3\times n_c}$ are
\begin{equation}
\mathbf{D}_0
=
\begin{bmatrix}
\mathbf{B}_{d}(0)\\
\mathbf{B}_{d-1}(0)\mathbf{D}^{(1)}\\
\mathbf{B}_{d-2}(0)\mathbf{D}^{(2)}
\end{bmatrix},
\qquad
\mathbf{D}_f
=
\begin{bmatrix}
\mathbf{B}_{d}(\tau)\\
\mathbf{B}_{d-1}(\tau)\mathbf{D}^{(1)}\\
\mathbf{B}_{d-2}(\tau)\mathbf{D}^{(2)}
\end{bmatrix}.
\end{equation}
Thus, reachable transition-state sets can be obtained by projecting the
feasible B-spline parameter set through these boundary mappings.

For a longer horizon $T$, we divide the trajectory into $N$ intervals
of duration $\tau=T/N$ and recursively propagate the reachable set
between successive intervals. For a linear mapping $\mathbf{E}$ and
polytope $\mathcal{P}$, we define
\begin{equation}
\operatorname{proj}_{\mathbf{E}}(\mathcal{P})
=
\left\{
\mathbf{E}\mathbf{p}
\mid
\mathbf{p}\in\mathcal{P}
\right\}.
\end{equation}
We initialize the recursion from the set of static states
\begin{align}
\mathcal{R}^{(0)}
&=
\left\{
\mathbf{s}=(q,0,0)
\mid
q_{\mathrm{min}}\leq q\leq q_{\mathrm{max}}
\right\}
\nonumber\\
&\equiv
\left\{
\mathbf{s}
\mid
\mathbf{G}^{(0)}\mathbf{s}\leq\mathbf{h}^{(0)}
\right\}.
\end{align}
Given a forward-reachable set
$\mathcal{R}_f^{(k)}
=
\{\mathbf{s}\mid
\mathbf{G}_f^{(k)}\mathbf{s}\leq\mathbf{h}_f^{(k)}\}$,
the set reachable after one additional interval is
\begin{align}
\mathcal{R}_f^{(k+1)}
&=
\left\{
\mathbf{s}_f
\middle|
\exists\,\mathbf{p}\in\mathcal{P}_{\mathrm{feasible}},
~
\mathbf{D}_0\mathbf{p}\in\mathcal{R}_f^{(k)},
~
\mathbf{D}_f\mathbf{p}=\mathbf{s}_f
\right\}
\nonumber\\
&=
\operatorname{proj}_{\mathbf{D}_f}
\left(
\left\{
\mathbf{p}
\middle|
\begin{bmatrix}
\mathbf{D}_c\\
\mathbf{G}_f^{(k)}\mathbf{D}_0
\end{bmatrix}
\mathbf{p}
\leq
\begin{bmatrix}
\mathbf{b}_c\\
\mathbf{h}_f^{(k)}
\end{bmatrix}
\right\}
\right)
\nonumber\\
&\equiv
\left\{
\mathbf{s}_f
\middle|
\mathbf{G}_f^{(k+1)}\mathbf{s}_f
\leq
\mathbf{h}_f^{(k+1)}
\right\},
\end{align}
with
$\mathbf{G}_f^{(0)}=\mathbf{G}^{(0)}$ and
$\mathbf{h}_f^{(0)}=\mathbf{h}^{(0)}$.
Repeating this propagation for $N$ intervals yields
\begin{equation}
\mathcal{R}_f^{(N)}
=
\left\{
\mathbf{s}
\mid
\mathbf{G}_f^{(N)}\mathbf{s}
\leq
\mathbf{h}_f^{(N)}
\right\}.
\end{equation}
The backward-reachable set
\begin{equation}
\mathcal{R}_b^{(N)}
=
\left\{
\mathbf{s}
\mid
\mathbf{G}_b^{(N)}\mathbf{s}
\leq
\mathbf{h}_b^{(N)}
\right\}
\end{equation}
is constructed analogously by initializing the terminal state with
$\mathcal{R}^{(0)}$ and propagating backward using the reversed boundary
mappings $\mathbf{D}_f$ and $\mathbf{D}_0$.

All polytope projections are computed in half-space representation
using the pypoman toolbox~\cite{caronpypoman}. Rather than parameterizing
the entire horizon $T$ with a single high-dimensional B-spline and
performing one projection, we recursively propagate the reachable sets
over shorter intervals using $n_c=6$ control points per interval.
Polyhedral projection in half-space representation becomes
computationally expensive and numerically unreliable as the dimension
of the parameter space increases. The recursive construction keeps each
individual projection low-dimensional while propagating reachability
over the full horizon.

The \textbf{Mutually Reachable Set} is obtained by intersecting the
forward- and backward-reachable sets:
\begin{align}
\mathcal{M}_S
&=
\mathcal{R}_f^{(N)}\cap\mathcal{R}_b^{(N)}
\nonumber\\
&=
\left\{
\mathbf{s}\in\mathcal{S}
\ \middle|\
\begin{bmatrix}
\mathbf{G}_f^{(N)}\\
\mathbf{G}_b^{(N)}
\end{bmatrix}
\mathbf{s}
\leq
\begin{bmatrix}
\mathbf{h}_f^{(N)}\\
\mathbf{h}_b^{(N)}
\end{bmatrix}
\right\}
\nonumber\\
&\equiv
\left\{
\mathbf{s}\in\mathcal{S}
\mid
\mathbf{M}\mathbf{s}\leq\mathbf{m}
\right\}.
\label{eq:mut_reach_Mm}
\end{align}

This intersection provides the mutual-reachability guarantee in
Eq.~\ref{eq:mut_reach_def}. Consider any two states
$\mathbf{s}_i,\mathbf{s}_f\in\mathcal{M}_S$. Because
$\mathbf{s}_i\in\mathcal{R}_b^{(N)}$, there exists a feasible trajectory
from $\mathbf{s}_i$ to the static-state set $\mathcal{R}^{(0)}$.
Likewise, because $\mathbf{s}_f\in\mathcal{R}_f^{(N)}$, there exists a
feasible trajectory from $\mathcal{R}^{(0)}$ to $\mathbf{s}_f$.
Concatenating these trajectories therefore provides a feasible connection
from $\mathbf{s}_i$ to $\mathbf{s}_f$. The reverse connection follows
in the same way because $\mathbf{s}_f\in\mathcal{R}_b^{(N)}$ and
$\mathbf{s}_i\in\mathcal{R}_f^{(N)}$. Thus, arbitrary states within
$\mathcal{M}_S$ can be connected in either direction while respecting
the prescribed physical limits.

The resulting polytope $\mathbf{M}\mathbf{s}\leq\mathbf{m}$ can be
imposed directly as a linear constraint in the online task-level
planner. The constructed $\mathcal{M}_S$ is an inner approximation of
the largest mutually reachable region because reachability is established
only over the finite-dimensional B-spline trajectory family used in the
offline construction. Feasible trajectories that cannot be represented
within this parameterization may therefore be excluded. Importantly,
this approximation affects the size of the resulting set rather than
the mutual-reachability guarantee: every state retained in
$\mathcal{M}_S$ satisfies the sufficient conditions established above.

The B-spline parameterization is used only for this offline construction
of $\mathcal{M}_S$; trajectories during online execution are generated
using Ruckig as described in the main text. The definition in
Eq.~\ref{eq:mut_reach_def} requires only that a feasible trajectory
exists between transition states, rather than that the trajectory
generated online use the same parameterization as the offline
construction.

Figure~\ref{fig:mut_reach} illustrates the resulting polyhedral sets for
the independent degrees of freedom using a horizon of
$T=1.0\,\mathrm{s}$, divided into $N=5$ intervals with $n_c=6$ control
points per interval.

\subsubsection*{Closed-form derivation of the regularized local linear dynamics}
We derive the analytic closed-form solutions $\mathbf{C}^*, \mathbf{D}^*,$ and $\mathbf{d}^*$ for the regularized weighted least-squares objective in \eqref{eq:reg_local_regression}. The objective function is given by:
\begin{align}
\mathcal{J}(\mathbf{C}, \mathbf{D}, \mathbf{d}) = &\sum_{i=1}^{k_{\text{NN}}} w_i \|\mathbf{y}_i - \mathbf{C} \delta\mathbf{x}_i - \mathbf{D} \delta\mathbf{u}_i - \mathbf{d}\|^2  \nonumber\\
&+ \gamma \left( \left\| \mathbf{C} - \mathbf{C}_0 \right\|_F^2 + \left\| \mathbf{D} - \mathbf{D}_0 \right\|_F^2 + \left\| \mathbf{d} - \mathbf{d}_0 \right\|^2 \right)
\end{align}
where $\delta\mathbf{x}_i = \mathbf{x}_i - \mathbf{x}$, $\delta\mathbf{u}_i = \mathbf{u}_i - \bar{\mathbf{u}}$, and prior model Jacobians are defined as $\mathbf{C}_0 \triangleq \frac{\partial \mathbf{f}_0}{\partial \mathbf{x}}(\mathbf{x}, \bar{\mathbf{u}})$, $\mathbf{D}_0 \triangleq \frac{\partial \mathbf{f}_0}{\partial \mathbf{u}}(\mathbf{x}, \bar{\mathbf{u}})$, and $\mathbf{d}_0 \triangleq \mathbf{f}_0(\mathbf{x}, \bar{\mathbf{u}})$. We construct block data matrices over the $k_{\text{NN}}$ nearest neighbors as:
\begin{align*}
\mathbf{Y} \quad\triangleq&\quad [\mathbf{y}_1 \, \cdots \, \mathbf{y}_{k_{\text{NN}}}] \\
\mathbf{Z} \quad\triangleq&\quad \begin{bmatrix} \delta\mathbf{x}_1 & \cdots & \delta\mathbf{x}_{k_{\text{NN}}} \\ \delta\mathbf{u}_1 & \cdots  & \delta\mathbf{u}_{k_{\text{NN}}} \\  1 & \cdots & 1\end{bmatrix} \\
\mathbf{W} \quad\triangleq&\quad \text{diag}(w_1, \dots, w_{k_{\text{NN}}})
\end{align*}
Concatenating parameters into $\mathbf{\Theta} \triangleq [\mathbf{C} \;\; \mathbf{D} \;\; \mathbf{d}]$ and prior terms into $\mathbf{\Theta}_0 \triangleq [\mathbf{C}_0 \;\; \mathbf{D}_0 \;\; \mathbf{d}_0]$, the objective simplifies in matrix notation to:
\begin{align*}
\mathcal{J}(\mathbf{\Theta}) ~=~ &\text{tr}\left( (\mathbf{Y} - \mathbf{\Theta}\mathbf{Z})\mathbf{W}(\mathbf{Y} - \mathbf{\Theta}\mathbf{Z})^\top \right)\\
&+ \gamma \, \text{tr}\left( (\mathbf{\Theta} - \mathbf{\Theta}_0)(\mathbf{\Theta} - \mathbf{\Theta}_0)^\top \right).
\end{align*}
Taking the matrix partial derivative with respect to $\mathbf{\Theta}$ yields $\frac{\partial \mathcal{J}}{\partial \mathbf{\Theta}} = -2\mathbf{Y}\mathbf{W}\mathbf{Z}^\top + 2\mathbf{\Theta}\left( \mathbf{Z}\mathbf{W}\mathbf{Z}^\top \right) + 2\gamma\mathbf{\Theta} - 2\gamma\mathbf{\Theta}_0$. Setting $\frac{\partial \mathcal{J}}{\partial \mathbf{\Theta}} = \mathbf{0}$ and recognizing that $\mathbf{Z}\mathbf{W}\mathbf{Z}^\top + \gamma \mathbf{I}$ is strictly positive definite for $\gamma > 0$, the optimal parameter matrix $\mathbf{\Theta}^*$ is obtained analytically as:
\begin{equation}
\label{eq:theta_star_cf}
\mathbf{\Theta}^* = [\mathbf{C}^* \;\; \mathbf{D}^* \;\; \mathbf{d}^*] = \left( \mathbf{Y}\mathbf{W}\mathbf{Z}^\top + \gamma \mathbf{\Theta}_0 \right) \left( \mathbf{Z}\mathbf{W}\mathbf{Z}^\top + \gamma \mathbf{I} \right)^{-1}.
\end{equation}

\subsection*{Related Work on Memory, Prior, and Adaptation}

Our work is related to robot-learning methods that use \emph{memory} of physical experience acquired during deployment to adapt behavior from prior knowledge. These approaches differ in how prior knowledge and accumulated experience are represented, and how that experience is retrieved or incorporated during online inference.  

In a broad sense, online reinforcement learning and offline-to-online reinforcement learning can also be viewed as forms of on-robot adaptation, since physical experience is incorporated into the model, value function, or policy during deployment \cite{ball2023efficient,dodeja2026q2rl}. We do not review this broader literature on general online policy improvement here, and instead focus on methods specifically designed for rapid adaptation from prior knowledge and limited physical experience. We broadly distinguish model-free approaches, which adapt behavior without explicitly performing inference through a dynamics model, from model-based approaches, which use experience to infer or update a model used for planning or control.  

Model-free approaches can acquire the structure for rapid adaptation during prior training, for example by conditioning policies on latent representations inferred from physical experience through direct optimization~\cite{yu2020learning}, probabilistic context inference~\cite{rakelly2019efficient}, or a pretrained adaptation network~\cite{kumar2021rma}. These approaches can adapt rapidly without explicit dynamics inference, although the available adaptation is shaped by the distribution of tasks and physical variations encountered during prior training. Memory of physical experience is typically represented through a finite recent history or incorporated implicitly into learned representations or parameters, rather than retained as an explicit growing memory from which relevant past experiences can later be retrieved.  

Recent sequence- and weight-level context models enable model-free policies to make use of longer-term memory more directly. LocoFormer~\cite{liu2025locoformer} extends Transformer context across trials, allowing attention to retrieve relevant information from previous interactions. This provides a learned alternative to the explicitly specified distance metric used for memory retrieval in our locally weighted formulation, though the computational and memory requirements of direct attention scale with context length. RoboTTT~\cite{jiang2026robottt} instead uses recurrent test-time parameter updates~\cite{zhang2026test,sun2024learning} to compress interaction history into model parameters, enabling longer-term memory without retaining the complete history within an attention context. 

Because our method instead combines explicitly stored physical experience with an explicit prior model, its closest methodological connections are to the model-based and nonparametric literature discussed next.  
PILCO~\cite{deisenroth2011pilco} is an early example of model-based learning from limited physical experience, using Gaussian Process (GP) regression to learn a probabilistic transition model and propagating the resulting uncertainty over multiple time steps for policy optimization. This provides a Bayesian nonparametric mechanism for combining accumulated observations with a prior over the dynamics. In its original formulation, however, PILCO uses a zero-mean GP prior, providing little guidance in regions of the state--action space where data have not yet been collected.  

Subsequent methods introduced more informative priors into this framework. Cutler and How~\cite{cutler2015efficient} construct a nonparametric dynamics prior from simulation and use it as the prior mean of a GP dynamics model, allowing physical observations to modify an informative simulated prediction. Chatzilygeroudis and Mouret~\cite{chatzilygeroudis2018using} similarly combine a parameterized black-box prior with a nonparametric model during model-based policy search. These approaches are particularly close to the motivation of our method: an imperfect prior can remain informative where physical experience is limited, while observations collected on the robot modify its prediction where evidence becomes available.  

There are nevertheless important differences in computational
structure. Standard GP regression involves inference with a covariance
matrix whose dimension grows with the number of retained observations
$N$, introducing an $\mathcal{O}(N^3)$ computational cost for exact
inference and motivating sparse or approximate GP methods as experience
accumulates. Our formulation instead retrieves a subset of
$k_{\mathrm{NN}}$ relevant experiences for each query and performs
local regression using only these samples. For a fixed feature
dimension, constructing the local regression problem scales linearly
with $k_{\mathrm{NN}}$, while the required matrix inversion has a fixed
dimension. Thus, the computational cost of local model fitting does not
grow directly with the total memory size $N$, allowing efficient online
inference as experience accumulates. The memory itself can nevertheless
continue to grow, motivating the memory-reduction strategies \cite{atkeson1997locally}.

The inference problems considered are also different. GP-based methods maintain uncertainty over learned dynamics and propagate it over multiple time steps for planning or policy optimization. In our experiments, the task decomposition instead provides a meaningful intermediate objective---the landing position of each throw---whose accurate execution supports the long-horizon juggling behavior. The learner therefore estimates the local command--outcome relationship required for the next throw, simplifying online inference and allowing each completed throw to become immediately available for subsequent adaptation. Locally weighted learning itself is not restricted to single-step inference: local dynamics models can also be used within multi-step control and planning formulations~\cite{atkeson1997locallycontrol}, albeit at additional computational cost.  

More recent approaches represent dynamics with neural networks and
meta-learn models that can be rapidly adapted from new experience.
Clavera et al.~\cite{clavera2018learning}, for example, use
MAML~\cite{finn2017model} to learn an initialization of a dynamics
model from which a small number of gradient updates can produce an
adapted model, followed by model predictive control. Prior knowledge is
therefore encoded implicitly in the meta-learned initialization, while
physical experience is incorporated through updates to a
fixed-dimensional set of model parameters without requiring all
previous observations to remain available at inference time. The
initialization and its response to gradient updates are learned from a
distribution of dynamics encountered during meta-training, enabling
rapid adaptation but making adaptation to qualitatively different model
mismatches dependent on generalization beyond this training
distribution. In contrast, nonparametric approaches such as GP
regression and our locally weighted formulation retain physical
observations explicitly and allow the locally inferred relationship to
change as new evidence accumulates, without requiring the particular
discrepancy to have been represented during prior training.

\begin{table*}[h]
    \centering
    \caption{\textbf{Optimization formulations of the task-level planner for throwing and catching tasks.} The planner maps high-level task commands $\mathbf{u}$ to target joint states $\mathbf{s}_f$.}
    \label{tab:sup_target_state_optimization}
    \begin{tabular}{l p{5.5cm} p{5.5cm}}
        \hline
        & \textbf{Throw} & \textbf{Catch} \\
        \hline
        \hline
        \textbf{Task-level command} ($\mathbf{u}$) & $\mathbf{p}^u_{\mathrm{land}}$ & $ \mathbf{p}_{\mathrm{catch}}$ \\
        \hline
        \textbf{Optimization variable} ($\mathbf{s}_f$) & $ (\mathbf{q}_f, \dot{\mathbf{q}}_f, \ddot{\mathbf{q}}_f)$ & $ (\mathbf{q}_f, \mathbf{0}, \mathbf{0})$ \\
        \hline
        \textbf{Objective} ($\phi$) & $\omega_p \lVert \mathbf{q}_f \rVert^2 + \omega_v \lVert \dot{\mathbf{q}}_f \rVert^2 + \omega_a \lVert \ddot{\mathbf{q}}_f \rVert^2$ & $\omega_p \lVert \mathbf{q}_f \rVert^2 + \omega_\theta (1 - \hat{\mathbf{n}}_{\mathrm{catch}}^\top \hat{\mathbf{n}}(\mathbf{q}_f))$ \\
        & $+ \omega_\theta (1 - \hat{\mathbf{n}}_{\mathrm{throw}}^\top \hat{\mathbf{n}}(\mathbf{q}_f))$ & \\
        \hline
        \textbf{Constraints} ($\mathbf{g}$) & $\mathbf{p}_{\mathrm{release}} = \mathbf{f}_{\mathrm{FK}}(\mathbf{q}_f)$ & $\mathbf{p}_{\mathrm{catch}} = \mathbf{f}_{\mathrm{FK}}(\mathbf{q}_f)$ \\
        & $\mathbf{v}_{\mathrm{throw}} = \mathbf{J}(\mathbf{q}_f)\dot{\mathbf{q}}_f$ &    \\
        & $\mathbf{a}_{\mathrm{throw}} = \mathbf{J}(\mathbf{q}_f)\ddot{\mathbf{q}}_f + \dot{\mathbf{J}}(\mathbf{q}_f, \dot{\mathbf{q}}_f)\dot{\mathbf{q}}_f$ & \\
        & $\hat{\mathbf{n}}_{\mathrm{throw}}^\top \hat{\mathbf{n}}(\mathbf{q}_f) \geq \cos(\theta_{\mathrm{max}})$ & $\hat{\mathbf{n}}_{\mathrm{catch}}^\top \hat{\mathbf{n}}(\mathbf{q}_f) \geq \cos(\theta_{\mathrm{max}})$ \\
        & $\mathbf{M} \mathbf{s}_f \leq \mathbf{m}$ \quad (see Equation \ref{eq:mut_reach_Mm}) & $\mathbf{M} \mathbf{s}_f \leq \mathbf{m}$ \quad (i.e., $\mathbf{s}_f \in \mathcal{M}_S$) \\
        \hline
        \hline
        \textbf{Fixed skill parameters} & $\mathbf{p}_{\mathrm{release}}$  &  \\
        \textbf{and relations} & $\mathbf{v}_{\mathrm{throw}} = \frac{1}{t_f}(\mathbf{p}^u_{\mathrm{land}} - \mathbf{p}_{\mathrm{release}}) + \frac{t_f}{2}\mathbf{g}$ & \\
        & $\mathbf{a}_{\mathrm{throw}} = \mathbf{g} = (0,~ 0,~ -9.81)$ &  \\
        & $\hat{\mathbf{n}}_{\mathrm{throw}} = \mathbf{v}_{\mathrm{throw}} / \lVert \mathbf{v}_{\mathrm{throw}} \rVert$ & $\hat{\mathbf{n}}_{\mathrm{catch}}$\\
        \hline
    \end{tabular}
\end{table*}

\begin{table*}[h]
	\centering
    \rowcolors{2}{white}{gray!15}
	\caption{\textbf{Skill sequence configuration for the cascade juggling pattern.} The Type parameter explicitly encompasses \texttt{throw} and \texttt{catch}. The Object ID specifies the tracked ball index ($0, 1, 2$), where $-1$ denotes an open-loop execution that bypasses visual tracking during a catch phase. The Task Identifier utilizes spatial targets ($R1, R2, L1, L2$) appended with specialized suffixes: \texttt{FIP} and \texttt{HIP} indicate fully opening or half-opening the index and pinky fingers while the thumb remains fully closed, whereas \texttt{Init} distinguishes introductory phase throws to facilitate memory-based learning. For skills within the stabilized cycle, the step duration is fixed at $0.35\text{ s}$, exactly half of the $0.7\text{ s}$ ballistic object flight time.}
	\label{tab:sup_pattern_cascade}

	\begin{tabular}{lll}
		\hline
		\textbf{Phase} & \textbf{Right Arm Skill} & \textbf{Left Arm Skill} \\
		\hline
		\textbf{Start} 
		& \texttt{[throw, 0, toL2fromR1FIP, 0.7]}       & \texttt{[catch, -1, atL2fromR1, 0.7]} \\
		& \texttt{[catch, -1, atR2fromL1, 0.3]}         & \texttt{[throw, 1, toR2fromL1Init, 0.35]} \\
		& \texttt{[throw, 2, toL2fromR1Init, 0.4]}      & \texttt{[catch, 0, atL2fromR1, 0.35]} \\
		\hline
		\textbf{Repeat} 
		& \multicolumn{2}{l}{\textit{Repeat the following sequence 5 times:}} \\
		& \texttt{[catch, 1, atR2fromL1, 0.35]}         & \texttt{[throw, 0, toR2fromL1, 0.35]} \\
		& \texttt{[throw, 1, toL2fromR1, 0.35]}         & \texttt{[catch, 2, atL2fromR1, 0.35]} \\
		& \texttt{[catch, 0, atR2fromL1, 0.35]}         & \texttt{[throw, 2, toR2fromL1, 0.35]} \\
		& \texttt{[throw, 0, toL2fromR1, 0.35]}         & \texttt{[catch, 1, atL2fromR1, 0.35]} \\
		& \texttt{[catch, 2, atR2fromL1, 0.35]}         & \texttt{[throw, 1, toR2fromL1, 0.35]} \\
		& \texttt{[throw, 2, toL2fromR1, 0.35]}         & \texttt{[catch, 0, atL2fromR1, 0.35]} \\
		\hline
		\textbf{Stop} 
		& \texttt{[catch, 1, atR2fromL1, 0.35]}         & \texttt{[throw, 0, toR2fromL1, 0.35]} \\
		& \texttt{[catch, -1, atR2fromL1FIP, 0.35]}     & \texttt{[catch, 2, atL2fromR1, 0.35]} \\
		& \texttt{[catch, 0, atR2fromL1HIP, 0.35]}      & \texttt{[catch, -1, atL2fromR1, 0.35]} \\
		\hline
	\end{tabular}
\end{table*}

\begin{table*}[h]
	\centering
    \rowcolors{2}{white}{gray!15}
	\caption{\textbf{Skill sequence configuration for tennis, half-shower, and cascade patterns.} The skill formatting (Type, Object ID, and Duration) remains consistent with Table~\ref{tab:sup_pattern_cascade}. The Task Identifier strings incorporate expanded spatial targets ($R3, R4, L3, L4$) alongside the cascade coordinates to accommodate the altered ballistic trajectories required across the distinct phases.}
	\label{tab:sup_pattern_tennis_halfshower_cascade}

	\begin{tabular}{lll}
		\hline
		\textbf{Phase} & \textbf{Right Arm Skill} & \textbf{Left Arm Skill} \\
		\hline
		\textbf{Start} 
		& \multicolumn{2}{l}{Identical to the cascade pattern start sequence} \\
		\hline
		\textbf{Repeat} 
		& \multicolumn{2}{l}{\textit{Phase A: Tennis Pattern. Repeat the following sequence 5 times:}} \\
		& \texttt{[catch, 1, atR2fromL1, 0.35]}         & \texttt{[throw, 0, toR3fromL4, 0.35]} \\
		& \texttt{[throw, 1, toL2fromR1, 0.35]}         & \texttt{[catch, 2, atL2fromR1, 0.35]} \\
		& \texttt{[catch, 0, atR3fromL4, 0.35]}         & \texttt{[throw, 2, toR2fromL1, 0.35]} \\
		& \texttt{[throw, 0, toL3fromR4, 0.35]}         & \texttt{[catch, 1, atL2fromR1, 0.35]} \\
		& \texttt{[catch, 2, atR2fromL1, 0.35]}         & \texttt{[throw, 1, toR2fromL1, 0.35]} \\
		& \texttt{[throw, 2, toL2fromR1, 0.35]}         & \texttt{[catch, 0, atL3fromR4, 0.35]} \\
		\cline{2-3}
		
		& \multicolumn{2}{l}{\textit{Phase B: Half-Shower Pattern. Repeat the following sequence 5 times:}} \\
		& \texttt{[catch, 1, atR2fromL1, 0.35]}         & \texttt{[throw, 0, toR2fromL1, 0.35]} \\
		& \texttt{[throw, 1, toL3fromR4, 0.35]}         & \texttt{[catch, 2, atL3fromR4, 0.35]} \\
		& \texttt{[catch, 0, atR2fromL1, 0.35]}         & \texttt{[throw, 2, toR2fromL1, 0.35]} \\
		& \texttt{[throw, 0, toL3fromR4, 0.35]}         & \texttt{[catch, 1, atL3fromR4, 0.35]} \\
		& \texttt{[catch, 2, atR2fromL1, 0.35]}         & \texttt{[throw, 1, toR2fromL1, 0.35]} \\
		& \texttt{[throw, 2, toL3fromR4, 0.35]}         & \texttt{[catch, 0, atL3fromR4, 0.35]} \\
		\cline{2-3}
		
		& \multicolumn{2}{l}{\textit{Phase C: Cascade Pattern. Repeat the following sequence 5 times:}} \\
		& \texttt{[catch, 1, atR2fromL1, 0.35]}         & \texttt{[throw, 0, toR2fromL1, 0.35]} \\
		& \texttt{[throw, 1, toL2fromR1, 0.35]}         & \texttt{[catch, 2, atL2fromR1, 0.35]} \\
		& \texttt{[catch, 0, atR2fromL1, 0.35]}         & \texttt{[throw, 2, toR2fromL1, 0.35]} \\
		& \texttt{[throw, 0, toL2fromR1, 0.35]}         & \texttt{[catch, 1, atL2fromR1, 0.35]} \\
		& \texttt{[catch, 2, atR2fromL1, 0.35]}         & \texttt{[throw, 1, toR2fromL1, 0.35]} \\
		& \texttt{[throw, 2, toL2fromR1, 0.35]}         & \texttt{[catch, 0, atL2fromR1, 0.35]} \\
		\hline
		\textbf{Stop} 
		& \multicolumn{2}{l}{Identical to the cascade pattern stop sequence} \\
		\hline
	\end{tabular}
\end{table*}

\begin{table*}[h]
	\centering
    \rowcolors{2}{white}{gray!15}
	\caption{\textbf{Skill sequence configuration for the shower juggling pattern.} The skill formatting (Type and Object ID) remains consistent with Table~\ref{tab:sup_pattern_cascade}. The Task Identifier utilizes spatial targets ($R5, L6$) appended with specialized behavior suffixes: \texttt{FIP} and \texttt{HIP} indicate fully opening or half-opening the index and pinky fingers while the thumb remains fully closed. The \texttt{Init} string distinguishes initial phase throws to facilitate memory-based learning. Within the stabilized cycle, the step duration is fixed at $0.24\text{ s}$, corresponding to the flight time of the lower horizontal pass and exactly one-third of the $0.72\text{ s}$ ballistic flight time of the higher arc throw.}
	\label{tab:sup_pattern_shower}

	\begin{tabular}{l l l}
		\hline
		\textbf{Phase} & \textbf{Right Arm Skill} & \textbf{Left Arm Skill} \\
		\hline
		\textbf{Start} 
		& \texttt{[throw, 0, toL6fromR5FIP, 0.7]}       & \texttt{[catch, -1, atL6fromR5, 0.35]} \\
		& \texttt{[catch, -1, atR5fromL6, 0.24]}        & \texttt{[catch, -1, atL6fromR5, 0.35]} \\
		& \texttt{[throw, 2, toL6fromR5Init, 0.24]}     & \texttt{[throw, 1, toR5fromL6Init, 0.48]} \\
		\hline
		\textbf{Repeat}
        & \multicolumn{2}{l}{\textit{Repeat the following sequence 5 times:}} \\
		& \texttt{[catch, -1, atR5fromL6, 0.24]}        & \texttt{[catch, 0, atL6fromR5, 0.24]} \\
		& \texttt{[throw, 1, toL6fromR5, 0.24]}         & \texttt{[throw, 0, toR5fromL6, 0.24]} \\
		& \texttt{[catch, -1, atR5fromL6, 0.24]}        & \texttt{[catch, 2, atL6fromR5, 0.24]} \\
		& \texttt{[throw, 0, toL6fromR5, 0.24]}         & \texttt{[throw, 2, toR5fromL6, 0.24]} \\
		& \texttt{[catch, -1, atR5fromL6, 0.24]}        & \texttt{[catch, 1, atL6fromR5, 0.24]} \\
		& \texttt{[throw, 2, toL6fromR5, 0.24]}         & \texttt{[throw, 1, toR5fromL6, 0.24]} \\
		\hline
		\textbf{Stop} 
		& \texttt{[catch, -1, atR5fromL6, 0.24]}        & \texttt{[catch, 0, atL6fromR5, 0.24]} \\
		& \texttt{[catch, -1, atR5fromL6, 0.24]}        & \texttt{[catch, -1, atL6fromR5FIP, 0.24]} \\
		& \texttt{[catch, -1, atR5fromL6, 0.24]}        & \texttt{[catch, 2, atL6fromR5FIP, 0.24]} \\
		\hline
	\end{tabular}
\end{table*}

\begin{table*}[h]
	\centering
    \rowcolors{2}{white}{gray!15}
	\caption{\textbf{Skill sequence configuration for the box juggling pattern.} The skill formatting (Type, Object ID, and Duration) remains consistent with Table~\ref{tab:sup_pattern_shower}. The Task Identifier extends spatial targets with $R6, L5$ and maintains the same special suffixes in Table~\ref{tab:sup_pattern_shower}.}
	\label{tab:sup_pattern_box}

	\begin{tabular}{l l l}
		\hline
		\textbf{Phase} & \textbf{Right Arm Skill} & \textbf{Left Arm Skill} \\
		\hline
		\textbf{Start} 
		& \texttt{[throw, 0, toL6fromR5FIP, 0.7]}       & \texttt{[catch, -1, atL6fromL5, 0.35]} \\
		& \texttt{[catch, -1, atR6fromR5, 0.24]}        & \texttt{[catch, -1, atL6fromL5, 0.35]} \\
		& \texttt{[throw, 2, toR6fromR5Init, 0.24]}     & \texttt{[throw, 1, toR5fromL6Init, 0.48]} \\
		& \texttt{[catch, -1, atR5fromL6, 0.24]}        & \texttt{[catch, 0, atL6fromR5, 0.24]} \\
		\hline
		\textbf{Repeat}
        & \multicolumn{2}{l}{\textit{Repeat the following sequence 5 times:}} \\
		& \texttt{[throw, 1, toL5fromR6, 0.24]}         & \texttt{[throw, 0, toL6fromL5, 0.24]} \\
		& \texttt{[catch, 2, atR6fromR5, 0.24]}         & \texttt{[catch, -1, atL5fromR6, 0.24]} \\
		& \texttt{[throw, 2, toR6fromR5, 0.24]}         & \texttt{[throw, 1, toR5fromL6, 0.24]} \\
		& \texttt{[catch, -1, atR5fromL6, 0.24]}        & \texttt{[catch, 0, atL6fromL5, 0.24]} \\
		\hline
		\textbf{Stop} 
		& \texttt{[catch, -1, atR6fromR5FIP, 0.24]}     & \texttt{[catch, -1, atL6fromL5, 0.24]} \\
		& \texttt{[catch, 2, atR6fromR5HIP, 0.24]}      & \texttt{[catch, -1, atL6fromL5, 0.24]} \\
		\hline
	\end{tabular}
\end{table*}


\clearpage 

\paragraph{Caption for Movie S1.}
\textbf{On-robot learning of the cascade juggling pattern.}
The movie shows the robot learning the three-ball cascade pattern from physical experience, with five consecutive cycles defined as successful completion. Starting from behavior generated using the analytical prior model, the robot updates its throwing commands from the outcomes of successive throws. The movie follows the progression from the initial unsuccessful attempts to successful execution of five consecutive cycles.

\paragraph{Caption for Movie S2.}
\textbf{On-robot learning across tennis, half-shower, and cascade juggling patterns.}
The movie shows a continuous learning experiment in which the robot sequentially learns the three-ball tennis, half-shower, and cascade patterns while retaining the physical experience accumulated in memory. The robot transitions between the three patterns without restarting the learning process, allowing previously acquired experience to be reused as new patterns are introduced. Five consecutive cycles of tennis, five cycles of half-shower, and five cycles of cascade are used as the respective completion criteria.

\paragraph{Caption for Movie S3.}
\textbf{On-robot learning of the shower juggling pattern.}
The movie shows the robot learning the three-ball shower pattern from physical experience, with five consecutive cycles defined as successful completion. Across successive attempts, the robot updates its throwing commands using observed ball landing positions and progressively improves its execution of the pattern.

\paragraph{Caption for Movie S4.}
\textbf{On-robot learning of the box juggling pattern.}
The movie shows the robot learning the three-ball box pattern from physical experience, with five consecutive cycles defined as successful completion. Across successive attempts, the robot updates the throwing commands for the different skills composing the pattern and progressively improves its execution of the three-ball box.




\clearpage 
\setcounter{figure}{0}

\makeatletter
\renewcommand{\fnum@figure}{Movie \thefigure}
\makeatother



\end{document}